\documentclass[10pt,twocolumn,letterpaper]{article}
\usepackage{cvpr}              

\definecolor{cvprblue}{rgb}{0.21,0.49,0.74}
\usepackage[pagebackref,breaklinks,colorlinks,allcolors=cvprblue]{hyperref}

\def\paperID{13631} 
\def\confName{CVPR}
\def\confYear{2025}

\usepackage{comment}
\usepackage[binary-units]{siunitx}
\usepackage{relsize}
\usepackage{ifthen}

\usepackage{graphics} 
\usepackage{rotating}
\usepackage{color}
\usepackage{enumerate}
\usepackage[T1]{fontenc}
\usepackage{psfrag}
\usepackage{epsfig} 
\usepackage{booktabs}
\usepackage{graphicx,url}
\usepackage{multirow}
\usepackage{array}
\usepackage{latexsym}
\usepackage{amsfonts}
\usepackage{amsmath}
\usepackage{amssymb}
\usepackage{xstring}
\usepackage{multirow}
\usepackage{xcolor}
\usepackage{prettyref}
\usepackage{flexisym}
\usepackage{bigdelim}
\usepackage{breqn} 
\usepackage{listings}
\let\labelindent\relax
\usepackage{enumitem}
\usepackage{xspace}
\usepackage{bm}
\graphicspath{{./figures/}}
\usepackage{tikz}
\usetikzlibrary{matrix,calc}

\usepackage{etex}
\usepackage{algpseudocode}
\usepackage{algorithm}
\usepackage{subcaption}
\usepackage{amsmath}

\usepackage{tabularx}
\usepackage{mathtools} 
\usepackage{amsthm}

\newrefformat{prob}{Problem\,\ref{#1}}
\newrefformat{def}{Definition\,\ref{#1}}
\newrefformat{sec}{Section\,\ref{#1}}
\newrefformat{sub}{Section\,\ref{#1}}
\newrefformat{prop}{Proposition\,\ref{#1}}
\newrefformat{app}{Appendix\,\ref{#1}}
\newrefformat{alg}{Algorithm\,\ref{#1}}
\newrefformat{cor}{Corollary\,\ref{#1}}
\newrefformat{thm}{Theorem\,\ref{#1}}
\newrefformat{lem}{Lemma\,\ref{#1}}
\newrefformat{fig}{Fig.\,\ref{#1}}
\newrefformat{tab}{Table\,\ref{#1}}

\newcommand{\bdmath}{\begin{dmath}}
\newcommand{\edmath}{\end{dmath}}
\newcommand{\beq}{\begin{equation}}
\newcommand{\eeq}{\end{equation}}
\newcommand{\bdm}{\begin{displaymath}}
\newcommand{\edm}{\end{displaymath}}
\newcommand{\bea}{\begin{eqnarray}}
\newcommand{\eea}{\end{eqnarray}}
\newcommand{\beal}{\beq \begin{array}{ll}}
\newcommand{\eeal}{\end{array} \eeq}
\newcommand{\beas}{\begin{eqnarray*}}
\newcommand{\eeas}{\end{eqnarray*}}
\newcommand{\ba}{\begin{array}}
\newcommand{\ea}{\end{array}}
\newcommand{\bit}{\begin{itemize}}
\newcommand{\eit}{\end{itemize}}
\newcommand{\ben}{\begin{enumerate}}
\newcommand{\een}{\end{enumerate}}

\newcommand{\myParagraph}[1]{{\bf #1.}\xspace}

\newcommand{\hide}[1]{}

\newcommand{\hiddenText}{{\color{gray} hidden text.}}
\newcommand{\hideWithText}[1]{\hiddenText}

\DeclareMathOperator*{\argmax}{arg\,max}

\newcommand{\prob}[1]{{\mathbb P}\left(#1\right)}

\newcommand{\blue}[1]{{\color{blue}#1}}

\newcommand{\linkToPdf}[1]{\href{#1}{\blue{(pdf)}}}
\newcommand{\linkToPpt}[1]{\href{#1}{\blue{(ppt)}}}
\newcommand{\linkToCode}[1]{\href{#1}{\blue{(code)}}}
\newcommand{\linkToWeb}[1]{\href{#1}{\blue{(web)}}}
\newcommand{\linkToVideo}[1]{\href{#1}{\blue{(video)}}}
\newcommand{\linkToMedia}[1]{\href{#1}{\blue{(media)}}}
\newcommand{\award}[1]{\xspace} 

\usepackage{float}
\usepackage{hyperref}
\usepackage{graphicx}
\usepackage{epstopdf}
\usepackage{epsfig}
\usepackage{amsmath}
\usepackage{xr}
\usepackage{framed}

\DeclareCaptionLabelSeparator{periodspace}{.\quad}
\newcommand{\sgraph}{\mathcal{S}}

\newcommand{\tasks}{\mathcal{T}}
\newcommand{\hib}{H-IB\xspace}

\newcommand{\name}{ASHiTA\xspace}
\newcommand{\hta}{HTA\xspace}

\newcommand{\primitives}{primitives layer\xspace}
\newcommand{\Primitives}{Primitives Layer\xspace}

\usepackage[normalem]{ulem}

\title{\LARGE \bf \name: Automatic Scene-grounded HIerarchical Task Analysis}
\author{
\normalsize{Yun Chang}\thanks{LIDS, Massachusetts Institute of Technology, MA, USA. {\tt\small \{yunchang, lcarlone\}@mit.edu}} \and
\normalsize{Leonor Fermoselle}\thanks{Robotics and AI Institute, MA, USA. {\tt\small \{lfermoselle, dta, jw\}@theaiinstitute.com}} \and
\normalsize{Duy Ta}\footnotemark[2] \and
\normalsize{Bernadette Bucher}\thanks{Department of Robotics, University of Michigan, MI, USA. {\tt\small bucherb@umich.edu}\\
This work was done partially while Chang was an intern at the Robotics and AI Institute and partially supported by the ARL DCIST Program.
} \and
\normalsize{Luca Carlone}\footnotemark[1] \and
\normalsize{Jiuguang Wang}\footnotemark[2]
}

\begin{document}

\thispagestyle{plain}
\pagestyle{plain}

\maketitle



\begin{abstract}
While recent work in scene reconstruction and understanding
has made strides in grounding natural language to physical 3D environments,
it is still challenging to ground abstract, high-level instructions to a 3D scene. 
High-level instructions might not explicitly invoke semantic elements in the scene, and even the process of breaking a high-level task into 
a set of more concrete subtasks ---a process called \emph{hierarchical task analysis}--- is environment-dependent. 
In this work, we propose \name,
the first framework that generates a task hierarchy grounded to a 3D scene graph
by breaking down high-level tasks into grounded subtasks.
\name alternates LLM-assisted hierarchical task analysis ---to generate the task breakdown---
with task-driven 3D scene graph construction to generate a suitable representation of the environment. 
Our experiments show that \name performs significantly better than
LLM baselines in breaking down high-level tasks into environment-dependent subtasks
and is additionally able to achieve grounding performance comparable to state-of-the-art methods.
\end{abstract}

\section{Introduction}
\label{sec:intro}


Modern machine vision applications, ranging from robotics to augmented reality, demand the development of 
vision systems that are able to parse and support the execution of high-level instructions, possibly provided by non-expert users.
For example, consider a robot that is given a high-level task of preparing for dinner.
To carry out this task,
the robot is required to decompose the task into granular subtasks
like fetching objects or visiting locations of interest (\eg, go to the kitchen, preheat the oven). 
Moreover, it must be able to ground these instructions into 3D objects and locations observed in the environment.

\begin{figure}[th]
\centering
\includegraphics[trim=0 0 0 0, clip, width=1.0\columnwidth]{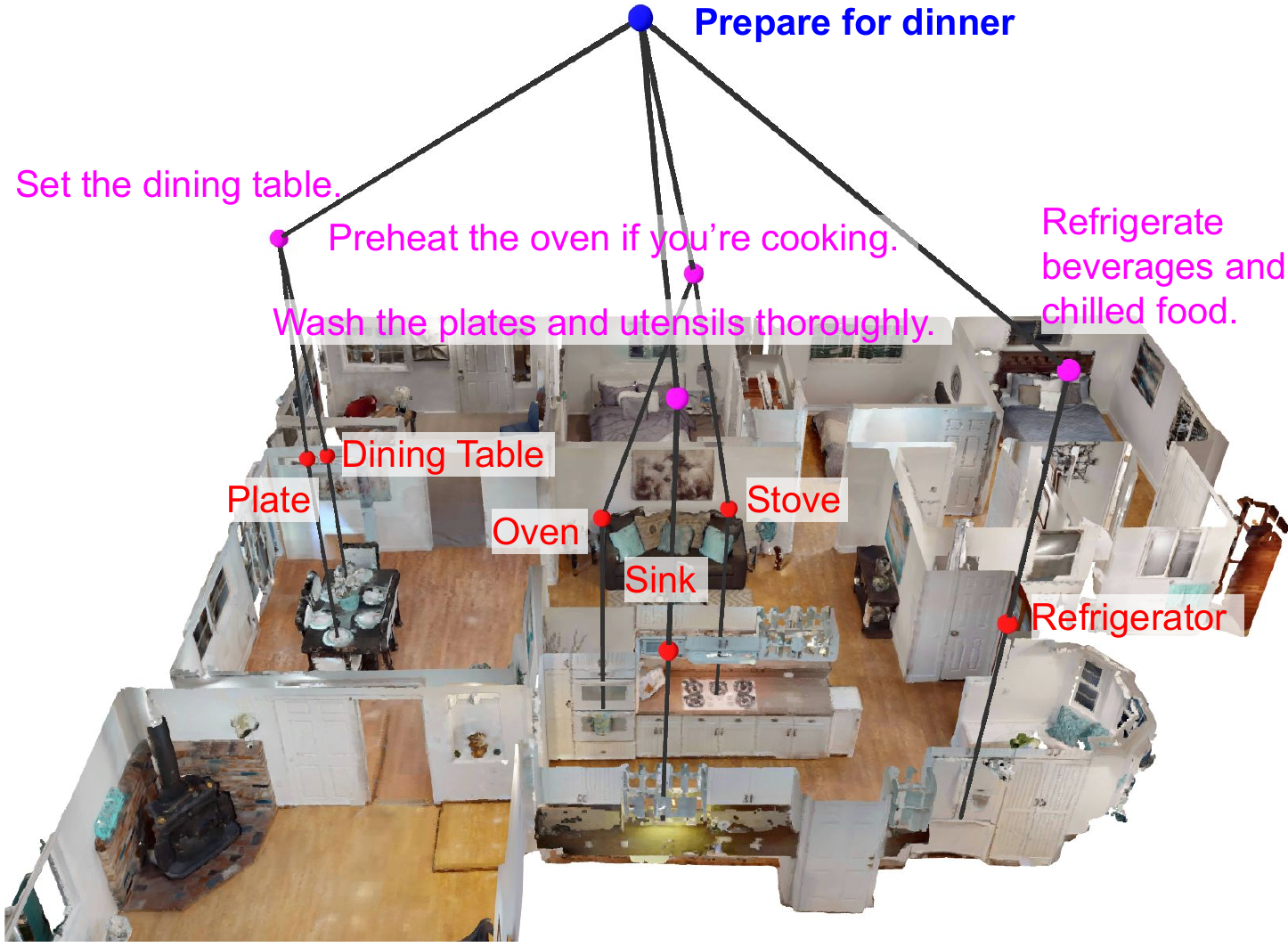}
\caption{
Given the high-level task of "Prepare for dinner",
\name automatically generates a hierarchy of subtasks (and items) while grounding them to a 3D scene graph.
For the scene graph in the figure, the blue node corresponds to the high-level task,
 magenta nodes correspond to the subtasks,
and red nodes correspond to the items required by the subtasks.
\vspace{-6mm}
\label{fig:cover}}
\end{figure}

Towards these goals, prior work has mostly focused on developing more expressive map representations, 
which combines both geometric and semantic information, to facilitate the grounding process.
Related work investigates
sparse object-based metric-semantic maps~\cite{Salas-Moreno13cvpr, Mo19iros-orcVIO},
dense metric-semantic maps~\cite{McCormac17icra-semanticFusion,Rosinol21ijrr-Kimera},
and 3D scene graphs~\cite{Armeni19iccv-3DsceneGraphs,Hughes24ijrr-hydraFoundations, Wu21cvpr-SceneGraphFusion}.
Traditionally, these representations have relied on \emph{closed-set semantic labels}, where the semantic concepts are restricted to the set of labels 
annotated in the training dataset.
 However, using a pre-defined set of labels limits the expressiveness of the map model (e.g., if our robot only has a ``chair'' class, it might not be able to distinguish a folding chair from a rocking chair), thus imposing constraints and creating ambiguity when grounding language commands.

To overcome this constraint, recent work has utilized new foundation models
 with \emph{open-set semantics} for mapping~\cite{Jatavallabhula23rss-ConceptFusion, Gu24icra-conceptgraphs, Werby24rss-hovsg}.
However, as pointed out in~\cite{Maggio24ral-clio}, 
when using open-set semantics, it becomes unclear how to group objects (e.g., is the zipper of a bag an object, or only the bag itself?)
Therefore, the work~\cite{Maggio24ral-clio} proposes an approach that clusters open-set concepts in the map in a task-driven fashion, thus representing the objects and places in a scene graph at the granularity required by specified natural-language tasks. 
However, the resulting approach, Clio~\cite{Maggio24ral-clio},
still cannot support high-level tasks and
is limited to supporting simple tasks that explicitly invoke relevant objects and locations (\eg, preheat the oven, go to the kitchen).
%
%

Other approaches to solving complex robotics tasks 
requiring semantic reasoning 
try to sidestep this task-to-visual grounding problem.
Side stepping this problem in tasks such as natural language navigation instruction following~\cite{Hong_2021_CVPR}, semantic search problems~\cite{Yokoyama24icra-vlfm, Chen23rss-HowToNot}, and natural-language-directed pick and place~\cite{Ahn22arxiv-sayCan, conceptgraphs} leads to effective problem-specific solutions.
More concretely, rather than rolling out the next action sequentially from what is currently seen by the robot’s camera~\cite{Ahn22arxiv-sayCan,Hong_2021_CVPR},
we use explicitly stored observations from the entire scene for task decomposition,
which enables the reasoning of tasks without clear termination conditions and over larger scenes.



{\bf Contribution.}
This paper presents the first framework to automatically generate a task hierarchy ---which breaks down high-level tasks specified in natural language into subtasks and items--- while grounding the task hierarchy into a 3D scene graph representation of the environment.

We start by considering the idealized case where the robot is not only given high-level tasks but also the grounded breakdown of the tasks into subtasks (what we call a \emph{task hierarchy}).
 Our first contribution is to generalize the Information Bottleneck (IB) principle~\cite{Tishby01accc-IB} used in~\cite{Maggio24ral-clio} to generate a \emph{hierarchy} of compressed representations: 
these are the layers of the 3D scene graph, which describe the scene at increasing levels of abstraction. 
We show that, under certain Markov assumptions between the random variables, 
we can derive an iterative algorithm for this hierarchical generalization of the IB (H-IB).
Given a task hierarchy, 
the H-IB creates a scene graph where the layers are arranged according to the task hierarchy. 

A fundamental issue is that in practice the robot is given the high-level task, but may not be given the task hierarchy (\ie might not be provided with the details of the subtasks to execute). This is a common situation since the task hierarchy itself is environment-dependent.
For instance, given the task ``clean the office'',
the exact breakdown and subtasks depend on the layout and items within the office in question.
Therefore, the key observation is that not only should the map representation depend on the task, but the way we break down the task depends on the information in our map representation.
Hence our second contribution is an iterative approach that alternates between 
a top-down hierarchical task analysis to generate a task hierarchy, 
and a bottom-up task-driven 3D scene graph construction to ground the task hierarchy. 
The result is \name, an approach for automatic scene-grounded hierarchical task analysis (Fig.~\ref{fig:cover}).
%

We evaluate \name on both the idealized case with given task hierarchies and also the case where only high-level tasks are specified.
Our experiments show that
(i) given a task hierarchy, \name is able to generate a task-driven 3D scene graph 
that is more accurate in grounding task-relevant objects than state-of-the-art zero-shot visual-grounding models~\cite{Zhu23cvpr-3DVisTA, Zhu24arxiv-Unifying3V}
and (ii) given high-level natural language tasks,
\name is able to automatically generate a task hierarchy grounded to a 3D scene graph.
To our knowledge, \name is the first system capable of accomplishing this,
and our experiments demonstrate competitive advantages over naive LLM and scene graph baselines.

\section{Related Work}
\label{sec:related_works}
{\bf{Scene Graphs.}}
Scene graphs have been used in computer graphics, vision, and robotics. 
2D scene graphs such as~\cite{Johnson15cvpr} have been used for image retrieval, generation, and understanding. 
3D scene graphs model 3D scenes using a graph where nodes are semantic concepts ---grounded in 3D--- and edges represent relations~\cite{Armeni19iccv-3DsceneGraphs,Kim19tc-3DsceneGraphs,Rosinol21ijrr-Kimera}. 
In contrast to voxel~\cite{shafiullah2022clip, liu2024DynaMem} or neural feature field~\cite{kerr2023lerf, shorinwa2024splat} based representations, 3D scene graphs provide structured, relational information that directly supports high-level reasoning and task planning by explicitly capturing object relationships and attributes.
SceneGraphFusion~\cite{Wu21cvpr-SceneGraphFusion} estimates a scene graph of objects and relations in real-time.
Hydra~\cite{Hughes22rss-hydra} is the first approach to build a hierarchical scene graph (including multiple layers) in real-time from sensor data.
These algorithms and models focus on a closed set of semantic labels, which fundamentally limits their ability to represent novel and open-ended concepts encountered in complex, real-world environments.


{\bf{Open-Set Semantic Understanding.}}
A number of recent works have explored the use of vision-language foundation models for open-set semantic understanding. ConceptGraph~\cite{Gu24icra-conceptgraphs} leverages vision-language models for 3D multi-view association, as well as SAM~\cite{Kirillov23iccv-SegmentAnything} and CLIP~\cite{Radford21icml-clip} to cluster a scene based on the objects' semantic and geometric similarity. HOV-SG~\cite{Werby24rss-hovsg}  leverages open-vocabulary vision foundation models to obtain segment-level maps in 3D and ultimately construct a hierarchy of concepts, though not in real time. 
CLIO~\cite{Maggio24ral-clio} builds a task-driven 3D scene graph, where an information-theoretic framework is used to select the subset of objects and scene structures that are relevant to the task specified in natural language.
While these approaches have made impressive strides towards semantically rich map representations, they still assume manually defined layers within the 3D scene graph,
thus constraining the type of language instructions that can be grounded~\cite{Maggio24ral-clio, Werby24rss-hovsg}.

{\bf{Hierarchical Task Analysis.}}
The breakdown of a complex long-horizon task into subtasks has been historically studied within the task and motion planning (TAMP) framework~\cite{zhao2024survey, garrett2021integrated}. Given a task in natural language, SayCan~\cite{Brohan23corl-saycan} uses LLMs to generate a set of feasible planning steps, re-scoring matched admissible actions with a learned value function, but limited by the size of the action space and the assumption of a 1 to 1 mapping between LLM output and action. LLM+P~\cite{liu23arxiv-llmp} uses an LLM as a PDDL writer in solving a planning problem described in natural language, prompted with a domain PDDL and example problem-PDDL pairs as context. Similarly, ProgPrompt~\cite{singh23icra-ProgPrompt} leverages a programmatic LLM prompt structure to generate an entire executable plan program directly, functional across situated environments, robot capabilities, and tasks. AutoTAMP~\cite{chen23arxiv-AutoTAMP} uses Signal Temporal Logic (STL), derived from state observation and language instructions, as the intermediary representation and performs autoregressive re-prompting to correct syntactic and semantic errors. 
To enable geometrically and kinematically feasible plans, recent works have explored the use of intermediate goals instead of actions \cite{yang2024guiding} and the generation of discrete and continuous language-parameterized constraints~\cite{kumar2024openworldtaskmotionplanning}.
The ability of LLMs to perform task planning remains under debate, with both positive~\cite{Silver23aaai-pddlLLM} and discouraging~\cite{valmeekam22neurips-PlanBenchAE} results.  Grounding the output of LLMs within the spatial and semantic structure of a scene graph~\cite{ray2024task} could potentially address the limitations of these approaches in their ability to integrate context-specific spatial information and actionable scene elements, which are essential for robust and contextually relevant task execution.



\section{Problem Formulation}
\label{sec:problem}

The core insight of \name is that when only given the visual observations of a 3D scene and high-level tasks,
the proper breakdown of the tasks into subtasks
depends on the available tools and objects in the scene representation,
while at the same time, a scene representation capable of supporting the task execution
depends on the task hierarchy.

\myParagraph{Hierarchical Task Analysis}
Hierarchical Task Analysis (\hta) is widely used in fields such as User Experience Design and Human-Computer Interaction~\cite{Norman02book-designEverydayThings, Diaper04-HandbookTaskAnalysis}.
It is used to break down complex tasks into a task hierarchy composed of atomic subtasks
to aid the understanding of how tasks are performed.
In the context of this work, our goal is to perform \hta of a task that is grounded in the physical scene
in order to understand how high-level tasks can be carried out with observed objects in a 3D environment.

\myParagraph{Task-Driven Scene Graph}
We construct a Task-Driven Scene Graph as a 
scene representation to ground the task hierarchy given by \hta.
Each node in the scene graph corresponds to a task,
a subtask, or an item involved in carrying out a subtask (Fig.~\ref{fig:cover}).
These nodes are organized hierarchically such that each task involves multiple subtasks,
and each subtask involves interactions with certain items.

\section{\name}
\label{sec:method}

\begin{figure*}[t]
\centering
\includegraphics[trim=0 0 0 0, clip, width=0.99\textwidth]{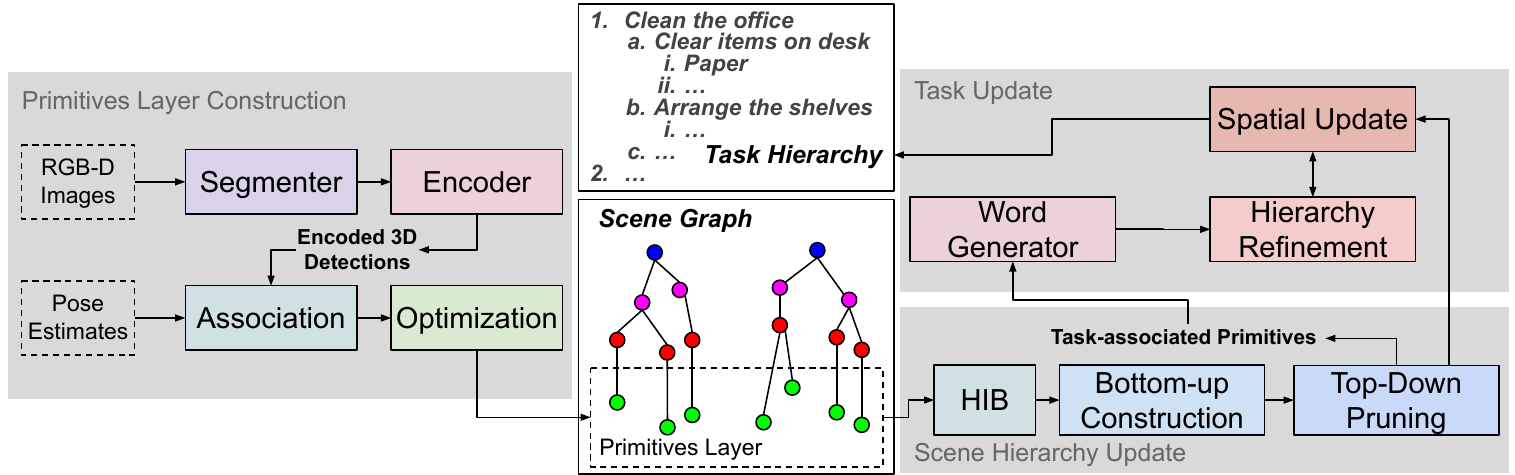}
\caption{\name first segments and encodes primitives in 2D, and then associates and optimizes them in 3D together with the camera poses. 
\name then breaks down high-level tasks into a task hierarchy
by alternating two steps:
a \emph{Scene Hierarchy Update} (Section~\ref{sec:scene_hierarchy}) which 
creates a 3D scene graph from the \primitives using the task hierarchy,
and a \emph{Task Update} (Section~\ref{sec:task_update}) which uses an LLM and the 3D scene graph to refine the task hierarchy.}
\label{fig:system}\vspace{-5mm}
\end{figure*}

This section describes \name, our approach for coupled \hta and task-driven 3D scene graph construction.
The architecture is summarized in Fig.~\ref{fig:system}.
\name consists of three main components:
(i) construction of the \primitives from visual input (Sec.~\ref{sec:primitives}),
(ii) scene hierarchy update (Sec.~\ref{sec:scene_hierarchy}),
where we construct and update the scene graph
based on the task hierarchy for a given high-level task,
and (iii) task update (Sec.~\ref{sec:task_update}),
where the task hierarchy is updated based on the generated scene graph.
Given a set of high-level tasks, \name starts with an 
initial task hierarchy
and iteratively performs scene hierarchy update and task update
to refine the task hierarchy and the scene graph in an alternating fashion.

\subsection{\Primitives Construction}
\label{sec:primitives}

The \primitives is the bottom layer of \name's task-driven scene graph.
It consists of a set of 3D bounding boxes with associated visual features.
To construct the \primitives from RGB-D inputs,
a frontend performs primitive detection and association,
and a backend jointly optimizes the estimated camera poses and primitive positions,
ensuring spatial and temporal consistency of the primitives.
The camera poses are queried from RTABMap~\cite{Labb18jfr-RTABMap}. 

The frontend relies on 
EfficientViT~\cite{Liu23cvpr-EfficientViT},
a class-agnostic segmentation model,
to detect class-agnostic 2D segments from the input RGB camera images
and MobileCLIP~\cite{AnasosaluVasu23axiv-MobileCLIP},
a visual-language encoder, to generate feature embeddings for each segment,
thus forming a set of 2D primitives detected in the input frame.
Note that the resolution of each primitive depends on the segmentation model,
which means that a primitive can contain any object or object part.
Each primitive is then projected to 3D and transformed to the world frame using the depth image
and the estimated camera pose, forming the 3D primitives. 
To associate the latest 3D primitives with previously seen primitives,
we first sample candidates from within a $1$m radius,
then formulate the problem of associating the latest primitives
$I=\{i_0, ... , i_N\}$ to the candidate primitives $J = \{j_0, ..., j_M\}$ 
as an assignment problem using an $N \times M$ score matrix,
where each entry contains the score $s(i, j)$ for matching primitive $i$ with $j$.
The score is defined as the weighted sum of visual similarity and semantic similarity between primitives $i$ and $j$
\begin{equation}
\label{similarity_score}
    s(i,j) = \omega_{text}\phi_{text}(i,j) + \omega_{visual}\sum_{bbox}k_{match}(i,j)
\end{equation}
The visual similarity term is the sum of the match scores~\cite{Lindenberger23iccv-LightGlue} of matched keypoints $k_{match}$~\cite{DeTone18cvpr-SuperPointSelfSupervised}
within the detection bounding box of primitives $i$ and $j$,
and the semantic similarity term is the cosine similarity $\phi_{text}$ of the feature embeddings of primitives $i$ and $j$.
We compute the optimal solution to this assignment problem with the Hungarian matching algorithm~\cite{Kuhn55nrl-HungarianAssignment},
by maximizing the total cumulative sum of scores for matching primitive $i$ with primitive $j$. 
After association, we optimize the camera poses and primitive positions using GTSAM~\cite{gtsam}.


\subsection{Scene Hierarchy Update}
\label{sec:scene_hierarchy}

To construct the scene hierarchy from the \primitives,
we use and extend a well-known tool from information theory, the \emph{Information Bottleneck}~\cite{Tishby01accc-IB}, which compresses 
a representation in a task-driven fashion.
In particular, we propose a hierarchical extension, dubbed the \emph{Hierarchical Information Bottleneck} algorithm (\hib), which compresses 
the primitives multiple times, according to the layers in the task hierarchy.
We use the output of \hib to first construct the scene graph in a bottom-up fashion,
then perform a top-down pass to refine and prune the parts of the scene graph that have low task relevance.

\myParagraph{Hierarchical Information Bottleneck}
The Information Bottleneck algorithm (IB)~\cite{Tishby01accc-IB} seeks 
to find a compressed representation $\sgraph$ of the input data $\sgraph_0$ that retains as much relevant information about the task $\tasks$
as possible while minimizing the amount of information about $\sgraph_0$ that is not useful for $\tasks$.
Intuitively, IB creates a "bottleneck" in the information flow between input $\sgraph_0$ and task $\tasks$,
and the compressed version of the input $\sgraph$
preserves the maximum amount of information about $\tasks$ while discarding irrelevant details from $\sgraph_0$.
Formally, this can be written as 
\begin{equation}\label{eq:ib}
    \min_{\prob{\sgraph | \sgraph_0}} I(\sgraph_{0} ; \sgraph) - \beta I(\tasks ; \sgraph)
\end{equation}
where $I(X ; Y)$ is the mutual information between $X$ and $Y$, and the parameter $\beta$ controls the amount of compression.

\hib is a generalization of IB to account for multi-resolution tasks $\tasks_1 \hdots \tasks_n$ and multiple levels of compression $\sgraph_1 \hdots \sgraph_n$.
Conceptually, this can be visualized as passing input information through a series of bottlenecks of varying sizes ---from large to small--- each representing different levels of abstractions.
Formally, we can write the new minimization functional as
\begin{equation}\label{eq:hib}
    \min_{\prob{\sgraph_{k} | \sgraph_{k-1}} \text{, } k = 1 \hdots n} \sum_{k=1}^n I(\sgraph_{k-1} ; \sgraph_k) - \beta \sum_{k=1}^n I(\tasks_k ; \sgraph_k)
\end{equation}

We solve for the minimum by taking the partial derivative of the Lagrangian of \eqref{eq:hib}
with respect to $\prob{s_k | s_{k-1}}$.
Assuming the Markov chain conditions $\tasks_n \hdots \leftarrow \tasks_1 \leftarrow \sgraph_0 \leftarrow \hdots \leftarrow \sgraph_n$,
we can write the Lagrangian in terms of known probabilities and $\prob{s_k | s_{k-1}}$.
Setting the derivative to zero
gives a set of multi-layer update steps
\begin{equation}\label{eq:hib_cases}
    \begin{cases}
        & p_\tau(s_k|s_{k-1}) = \frac{1}{\mathcal{Z}} p_\tau(s_k) \exp(-\beta d) \\
        & p_{\tau+1}(s_k) = \sum_{s_{k-1}}p_\tau(s_{k-1})p_\tau(s_k | s_{k-1}) \\
        & p_{\tau+1}(t_k|s_k) = \sum_{s_{k-1}}p_\tau(t_k|s_{k-1})p_\tau(s_{k-1}|s_k)
    \end{cases}
\end{equation}
where $d$ is a weighted sum of the Kullback–Leibler Divergence $D_{KL}$
across the multi-resolution tasks.
\begin{equation}
\begin{split}
    d = & D_{KL}(p_\tau(t_k|s_k) || p_\tau(t_k|s_{k-1}))\\
    +& \sum_{i=k+1}^n\sum_{s_i}p_\tau(s_i|s_k)D_{KL}(p_\tau(t_i|s_i) || p_\tau(t_i|s_{k-1}))
\end{split}
\end{equation}
Note that we recover the standard IB~\cite{Tishby01accc-IB} when $n = 1$.
The full derivation is given in the supplementary material.

As mutual information is always non-negative and $I(\tasks_k ; \sgraph_k)$ is bounded by the entropy $H(\sgraph_k)$,
the functional \eqref{eq:hib} is bounded from below, preventing it from decreasing indefinitely.
Let us call $\mathcal{C}_\tau$ the objective function value at iteration $\tau$ in~\eqref{eq:hib}.
Each iteration step is independent and reduces the objective with respect to one of the distributions $p(s_k|s_{k-1})$, $p(s_k)$, and $p(t_k|s_k)$ while keeping the others fixed, therefore $\mathcal{C}_{\tau+1} \leq \mathcal{C}_\tau$.
Since $\mathcal{C}_\tau$ is non-increasing and bounded below, the sequence converges.
We remark that since the objective is not jointly convex in all distributions simultaneously, 
the algorithm may converge to different local minima depending on the initialization.



\begin{figure}[t]
    \centering
    \begin{subfigure}[b]{0.49\columnwidth}
        \centering
        \includegraphics[width=\linewidth]{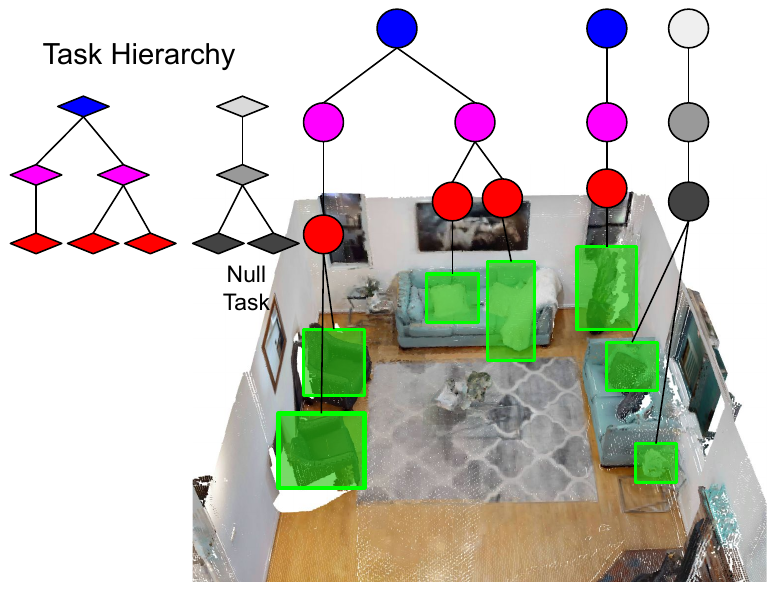} 
        \caption{Bottom-Up Construction}
        \label{fig:bottom_up_build}
    \end{subfigure}
    \hfill
    \begin{subfigure}[b]{0.49\columnwidth}
        \centering
        \includegraphics[width=\linewidth]{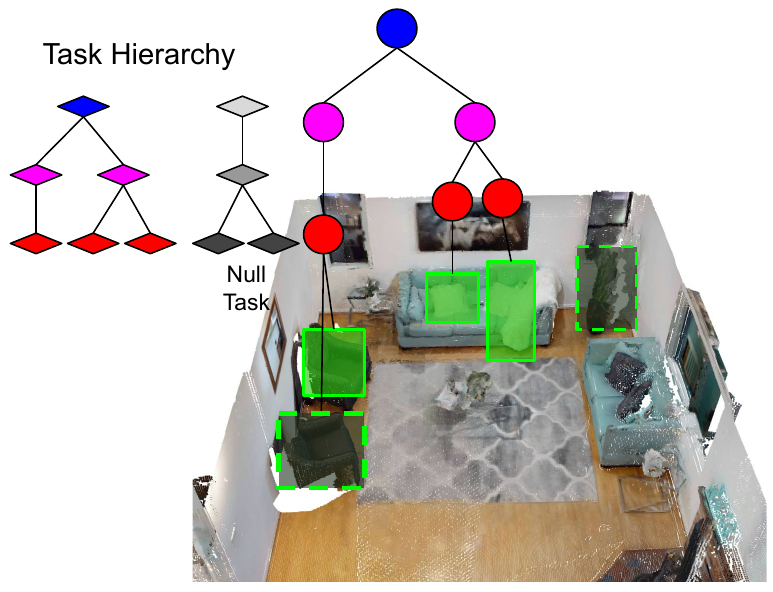} 
        \caption{Top-Down Pruning}
        \label{fig:top_down_prune}
    \end{subfigure}
    
    \vskip\baselineskip 
    
    \begin{subfigure}[b]{0.49\columnwidth}
        \centering
        \includegraphics[width=\linewidth]{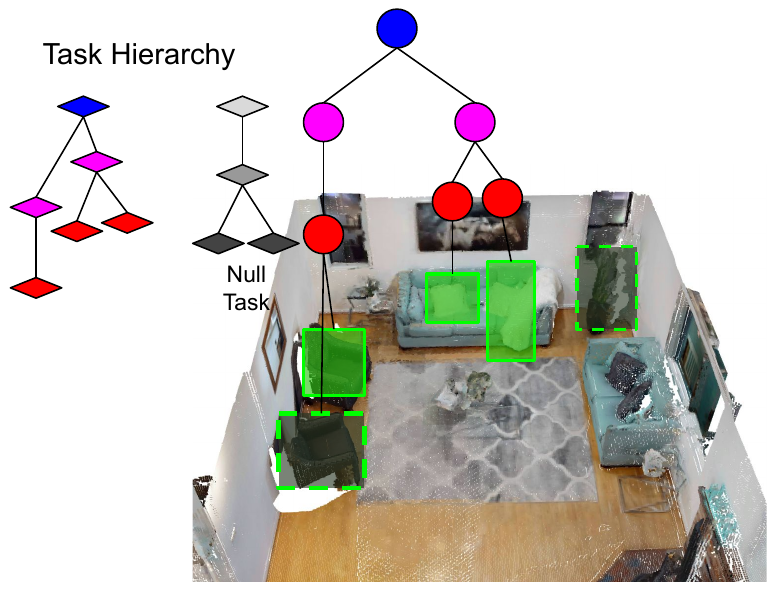}
        \caption{Task Spatial Update}
        \label{fig:spatial_update}
    \end{subfigure}
    \hfill
    \begin{subfigure}[b]{0.49\columnwidth}
        \centering
        \includegraphics[width=\linewidth]{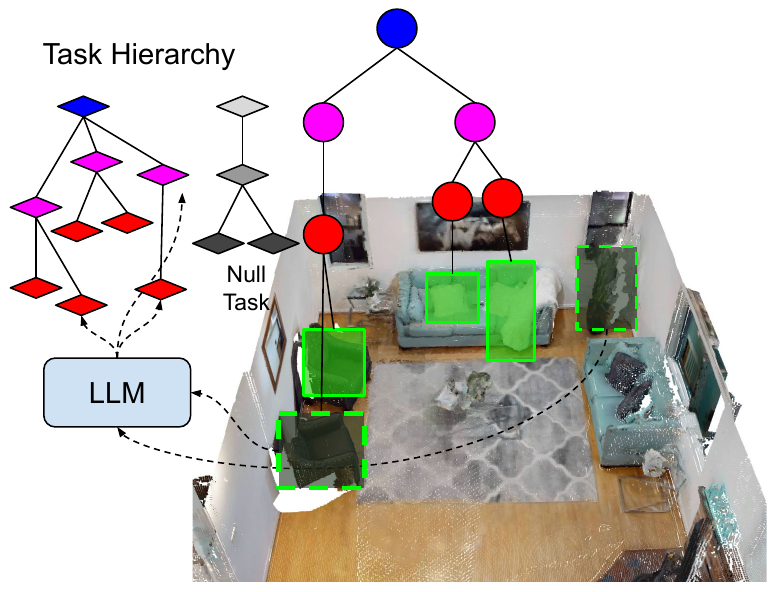} 
        \caption{Task Hierarchy Refinement}
        \label{fig:hierarchy_refinement}
    \end{subfigure}
    
    \caption{\name's Scene Hierarchy and Task Update steps.
    The task hierarchy is on the left with diamond-shaped nodes representing the task entities. 
    The scene graph is on the right, with circles marking the task-aligned scene graph nodes and green boxes marking the \primitives.
    (a) Bottom-Up Construction: Starting from an initial task hierarchy, we perform \hib and use the result to construct a 3D scene graph.
    (b) Top-Down Pruning: We perform pruning using the probabilities obtained from \hib and also prune nodes related to the null tasks.
    (c) Spatial Update: Using the scene graph, we can update the spatial locations of the tasks, subtasks, and items.
    (d) Hierarchy Refinement: With the suggested items given by \hib, we query the LLM to refine the task hierarchy.}
    \label{fig:ashita_sketch}
    \vspace{-5mm}
\end{figure}

\myParagraph{Bottom-Up Construction}
Starting with an ungrounded initial task hierarchy,
we generate a rough version of the scene graph using \hib as sketched out in Fig.~\ref{fig:bottom_up_build}.
Starting with the \primitives,
we compute the normalized cosine similarities between the primitives (green boxes in Fig.~\ref{fig:bottom_up_build}) and the text embedding of the items in the task hierarchy (red diamonds in Fig.~\ref{fig:bottom_up_build})
and take all primitives that have similarity of greater than $0.8$ to form the input $\sgraph_0$ to \hib.
The multi-level abstraction of the tasks corresponds to the task hierarchy
such that $\tasks_1$ corresponds to the items in subtasks, $\tasks_2$ corresponds to the subtasks,
and $\tasks_3$ corresponds to the high-level tasks.
To account for the irrelevant parts of the scene, we augment our task hierarchy with a \emph{null} task
that consists of a single null action step with items: "item" and "thing"~\cite{Kerr23iccv-lerf},
shown as the gray diamonds in Fig.~\ref{fig:bottom_up_build}.

\hib takes as input the conditional probabilities $\prob{\tasks_1|\sgraph_0}$, $\prob{\tasks_2|\sgraph_0}$, and $\prob{\tasks_3|\sgraph_0}$.
We compute these using the text embeddings of the items in the task hierarchy
and the embeddings of the nodes in the \primitives~\cite{Maggio24ral-clio}.
For $\tasks_1$, we compute the cosine-similarities scores between each embedding $f_t$, $t \in \tasks_1$ 
and embedding $f_s$, $s \in \sgraph_0$ and use the softmax function to obtain $\prob{\tasks_1|\sgraph_0}$.
From this, we can compute the conditional probability of the subtasks $\prob{\tasks_2|\sgraph_0}$ for $t_2\in \tasks_2$ and $s \in \sgraph_0$.
\begin{equation}
    p(t_2|s) = \sum_{t_1 \in \tasks_1}p(t_2|t_1)p(t_1|s)
\end{equation}
and the conditional probability of the tasks $\prob{\tasks_3|\sgraph_0}$
\begin{equation}
    p(t_3|s) = \sum_{t_2 \in \tasks_2}p(t_3|t_2)p(t_2|s)
\end{equation}

The conditional probabilities are computed bottom-up and do not require the sentence embeddings for the subtask or task descriptions.
This is in-line with the Markov chain assumption made when deriving for the \hib  iterations.

We solve \hib with $\beta = 10$ and with a min iteration of 10 and a max iteration of 1000 or until convergence, defined as $\mathcal{C}_\tau - \mathcal{C}_{\tau+1} < 10^{-8}$.
The output \hib are the probabilities $\prob{\sgraph_1|\sgraph_0}$, $\prob{\sgraph_2|\sgraph_1}$, and $\prob{\sgraph_3|\sgraph_2}$.
Each conditional probability encodes a soft mapping from layer $i$ to layer $i+1$ (\ie, the probability that an element in layer $i$ is a child of an element in layer $i+1)$.
We take the $\argmax$ of the conditional probabilities to
obtain the inter-layer edges, corresponding to the highest probability mapping,
between the \primitives to the items,
the items to the subtasks, and the subtasks to the tasks, respectively.
We also map scene graph nodes to the highest probability task, subtask, or item in the task hierarchy
by taking the $\argmax$ of $\prob{\tasks_k | \sgraph_k}$.
Intuitively, this step takes a set of primitives
and funnel them through the task hierarchy to obtain a 3D scene graph that is aligned with the task hierarchy (Fig.~\ref{fig:bottom_up_build}).

\myParagraph{Top-Down Pruning}
In some cases,
\hib produces multiple nodes associated to the same task entity (\ie, task, subtask, or item).
In this step, we first merge any overlapping scene graph nodes associated to the same task entity and then discard irrelevant nodes top-down,
as shown in Fig.~\ref{fig:top_down_prune}.
To do this,
we define \emph{confidence} as the probability of a node $s$ given the task entity $t$: $p(s|t)$.
This can be easily computed from $p(t|s)$, obtained from \hib,
along with $p(s)$ and $p(t)$.
Starting from the top of the task hierarchy, for each task, we compare all the associated nodes of the task and only keep the node with the highest confidence, pruning away all other nodes and their descendants (\eg the deleted blue circles and its subtask and item nodes in Fig.~\ref{fig:top_down_prune}).
We repeat this for the subtasks, and then also the items.
Additionally, we prune all nodes and their descendants that are associated with the \emph{null} task (\eg deleted gray circles in Fig.~\ref{fig:top_down_prune}),
as they are considered irrelevant to the given tasks.
Lastly, we prune the primitives such that for each scene graph node on the items layer,
we retain only the primitive with the highest confidence along with the primitives that intersect this highest confidence primitive.
After pruning, we assign a bounding box and centroid to the item-associated scene graph node based on the union of the remaining primitives.

\subsection{Task Update}
\label{sec:task_update}
The task hierarchy has to be both \emph{spatially localized} (i.e., each subtask should be executed at a location in the map),
and \emph{environment-aware} (i.e., the task should reference elements in the environment).
Hence, our task update consists of two steps: spatial update and hierarchy refinement.

\myParagraph{Spatial Update}
Initially, the task entities are not grounded,
hence do not have any spatial information.
Once a task entity in the task hierarchy is aligned with a scene graph node,
the task entity is \emph{grounded},
and we update the spatial information of the task entity and its descendants
accordingly, as sketched out in Fig.~\ref{fig:spatial_update}.
The position of a task entity is the centroid of the aligned scene graph node.
The radius of a task entity is defined as the distance from the node to its nearest neighbor for the tasks and subtasks,
or the Euclidean norm of its bounding box dimension for the items.
This spatial approximation is accounted for in the input conditional probabilities
for the next round for \hib,
as we will discuss below.

\myParagraph{Hierarchy Refinement}
The purpose of this step is to update the subtasks and items in the task hierarchy to
account for primitives that are not fully aligned to the task hierarchy during Scene Hierarchy Update (Sec.~\ref{sec:scene_hierarchy}),
shown as the dark green boxes in Fig.~\ref{fig:hierarchy_refinement}.
This is done by tracking the subtasks and the primitives that are assigned to the subtasks during
bottom-up construction (Fig.~\ref{fig:bottom_up_build}),
but are pruned during the top-down step (Fig.~\ref{fig:top_down_prune}).
Recall that each primitive includes a visual embedding (Sec.~\ref{sec:primitives}). 
The Word Generator takes in these primitives and passes them through a word bank
---an LLM-generated list of household items--- and finds the items with the highest CLIP similarities.
The output is a list of items present in the environment that can be used for each subtask.
For each subtask, we query an LLM (\texttt{Chat-GPT-4o-mini}) to come up with a score between 0 and 1 for each item on the suggested list.

\begin{framed}
\begin{quote}
    "Give me the probability values of each item in the list <suggestion> being
    required for the action: <subtask>.
\end{quote}
\end{framed}
We add the items with a score higher than a defined threshold $r_s$ to the subtask, otherwise the suggested item is stored to query for new subtasks.
Then, for each task,
we collect the remaining suggested items not incorporated by its subtasks,
and query an LLM for the likelihood scores in a similar manner, but with respect to the task.
We collect the items with scores above a defined threshold $r_t$
and query the LLM for additional subtasks as follows,
\begin{framed}
\begin{quote}
    "Given the previous steps for <task>,
    add additional steps that involves only the following items: <suggested-items>.
    Make sure that the same item does not appear in more than one step."
\end{quote}
\end{framed}
The new subtasks are then added to the task hierarchy and incorporated in the next scene hierarchy update.
This process is shown in Fig.~\ref{fig:hierarchy_refinement}.
In our experiments, we set $r_s = r_t = 0.8$ as the relevance threshold.

\myParagraph{Spatially-Informed Conditional Probability}
After task entities are grounded, they include spatial attributes,
which need to be accounted for when computing the conditional probabilities for \hib (Sec.~\ref{sec:scene_hierarchy}).
We define a spatial conditional probability for primitive $s_i$ and task entity $t$,
\begin{equation}
    p_s(s_i | t) = \eta \begin{cases} 1 & d < r\\
                    \exp{(- (d - r)^2 / r^2)} & d \geq r;
    \end{cases}
\end{equation}
where $\eta$ is a normalization constant, $d$ is the distance of a primitive to the item task entity,
and $r$ is the radius of the spatial attribute.
We can then compute
\begin{equation}
    p_s(t_k|s_i) = p_s(s_i | t_k) p(t_k) / p(s_i)
\end{equation}
and define the new distribution as 
\begin{equation}
    p(t_k|s_i) =  p_s(t_k|s_i) p_e(t_k|s_i) / \sum_k (p_s(t_k|s_i) p_e(t_k|s_i))
\end{equation}
where $p_e$ is the embeddings-based conditional probability discussed in Sec.~\ref{sec:scene_hierarchy}.
These distributions are then used in the next round of the \hib.

\section{Experiments}
\label{sec:experiment}

In this section, we first evaluate the more straightforward case where the high-level tasks are already broken down into subtasks.
We verify that \name is able to ground these given tasks and subtasks to objects in the scene (Sec.~\ref{sec:grounding_exp}).
Next, we evaluate the main contribution of this paper:
the automatic generation of a task hierarchy grounded in a 3D scene graph given a high-level task (Sec.~\ref{sec:task_exp}).
To justify the design choices made in \name,
we also perform an ablation study on the various components of \name (Sec.~\ref{sec:ablation}).
Lastly, we qualitatively demonstrate the performance of \name in real-world scenes given various high-level tasks (Sec.~\ref{sec:qualitative}).

\subsection{Grounding Evaluation}\label{sec:grounding_exp}

\begin{table}%
    \small
    \centering
    \begin{tabular}{c cc}
    \toprule
    Method & s-acc (\%) & t-acc (\%)\\
    \midrule
    3D-VisTA~\cite{Zhu23cvpr-3DVisTA} & 25.3 & 10.3 \\
    PQ3D~\cite{Zhu24arxiv-Unifying3V} & 24.4 & 9.7 \\
    ASHiTA & \bf{28.71} & \bf{12.13} \\
    \rowcolor{blue!10}
    ASHiTA + Txt Emb. & 65.4 & 39.33 \\
    \rowcolor{blue!10}
    GPT w/ GT labels~\cite{zhang24arxiv-sg3d} & \bf{75.9} & \bf{52.1} \\
    \midrule 
    \end{tabular}
    \caption{Evaluation of sequential grounding using the SG3D HM3DSem dataset with ground truth 3D instance segmentation.
    Trials with knowledge of the ground truth object labels are highlighted.
    \label{tab:grounding_gt}}\vspace{-5mm}
\end{table}

\begin{table}%
    \small
    \centering
    \begin{tabular}{c cc}
    \toprule
    Method & s-acc (\%) & t-acc (\%)\\
    \midrule
    Hydra~\cite{Hughes22rss-hydra} + GPT & 8.18 & 2.44 \\
    HOV-SG~\cite{Werby24rss-hovsg} & 8.98 & 1.95 \\
    ASHiTA & \bf{21.7} & \bf{8.78} \\
    \rowcolor{blue!10}
    Hydra (GT Seg) + GPT & 14.2 & 6.34 \\
    \midrule
    \end{tabular}
    \caption{Evaluation of sequential grounding on 8 scenes from the SG3D HM3DSem dataset with incrementally built scene-graph representations. Trials with knowledge of the ground truth object labels are highlighted.
    \label{tab:grounding}}\vspace{-6mm}
\end{table}

We evaluate \name's ability to ground subtasks to objects in the scene using the HM3DSem~\cite{Yadav22-habitatchallenge}
test scenes with annotated task decomposition and grounding given by the SG3D dataset~\cite{zhang24arxiv-sg3d},
which contains 890 high-level tasks distributed across 35 HM3DSem scenes.

To adapt \name for this evaluation, we use the given tasks and subtasks and generate an item for
each subtask with \texttt{Chat-GPT-4o-mini} to form the initial task hierarchy.
We only run half of the Hierarchy Refinement described in Sec.~\ref{sec:task_update} to only update the items for each subtask
and omit the step to add additional new subtasks.

We first evaluate the performance when given the ground truth 3D instance segmentation.
We compare the performance of \name with image embeddings and \name with the text embeddings of the ground-truth labels against
state-of-the-art zero-shot 3D visual grounding models (3D-VisTA~\cite{Zhu23cvpr-3DVisTA} and PQ3D~\cite{Zhu24arxiv-Unifying3V})
as benchmarked in~\cite{zhang24arxiv-sg3d}.
We use two metrics in our evaluation~\cite{zhang24arxiv-sg3d}: subtask-accuracy (s-acc) and task-accuracy (t-acc) for given regions of the scene.
The subtask-accuracy is the percentage of subtasks that are grounded to the correct object,
and the task-accuracy is the percentage of tasks with subtasks that are all correctly grounded.
The results are shown in Table~\ref{tab:grounding_gt}.
\name's performance is significantly better than the baseline zero-shot models.
The variants with ground truth information demonstrates performance without segmentation noise and with ideal visual-language alignment.
In this limit, \name surpasses the best fine-tuned baseline in SG3D~\cite{zhang24arxiv-sg3d}
and approaches the performance of GPT with ground-truth labels -- falling short due to the lack of explicit relational information (\eg ``next to'', ``on top of'') in \name.

To evaluate approaches better suited for embodied applications that incrementally build scene graphs,
we also use 8 RGB-D sequences generated by \cite{Werby24rss-hovsg} from 8 HM3DSem test scenes. 
This corresponds to 205 high-level tasks from the SG3D dataset~\cite{zhang24arxiv-sg3d}.
Our baselines for this experiment are to use GPT to ground the task hierarchy on Hydra~\cite{Hughes24ijrr-hydraFoundations} and HOV-SG~\cite{Werby24rss-hovsg} generated scene graphs.
For Hydra, we evaluate with both ground-truth 2D instance segmentation and with a closed-set segmentation model (OneFormer~\cite{Jain23cvpr-Oneformer}),
crop the resulting scene graph to the given evaluation region, then convert the resulting scene graph to a query-able format: 
\{Room1: \{Object 1: \{label: <object-name>, positions: (x, y, z)\}, Object2: {...}\}, Rom2: \{\}\}.
This is given to \texttt{GPT-4o-mini} alongside the tasks and the subtasks, and GPT identifies the node associated to each subtask.
For HOV-SG, we also first crop the scene graph of the full scene to the given region,
then we query the scene graph with the subtasks as described in~\cite{Werby24rss-hovsg} using \texttt{GPT-4o-mini} 
to retrieve the grounded object for each subtask.
For evaluation, we require the centroid of the estimated object bounding box to be contained by the ground-truth object bounding box to be considered a correct match.
The results are shown in Table~\ref{tab:grounding}.
We show that \name significantly outperforms the GPT-powered baselines,
even when ground-truth segmentation is used in Hydra, \cf Hydra (GT Seg) in the table.

\subsection{Hierarchical Task Analysis}\label{sec:task_exp}
We adapted the SG3D dataset~\cite{zhang24arxiv-sg3d} to evaluate hierarchical task analysis 
using the given subtasks as the reference task hierarchy for each high-level task.
To evaluate a generated task hierarchy, 
we count an estimated subtask as \emph{correct} if it grounds to the same set of objects as a reference subtask and \emph{incorrect} otherwise.
A reference subtask is \emph{missed} if there is no estimated subtask that grounds to its set of objects.
We report three metrics: subtask recall, step precision, and task accuracy.
Given $C$ correct subtasks, $I$ incorrect subtasks, and $M$ missed reference subtasks,
the subtask recall (s-rec) is defined as $\frac{C}{C + M}$, and
the subtask precision (s-prec) is defined as $\frac{C}{C + I}$.
The task accuracy (t-acc) is the percentage of tasks with no incorrect subtasks.

We utilize the 8 RGB-D sequences~\cite{Werby24rss-hovsg} and the associated 205 high-level tasks for evaluation.
We again use \texttt{GPT-4o-mini} and the scene graphs generated by Hydra~\cite{Hughes22rss-hydra}
and HOV-SG~\cite{Werby24rss-hovsg} as baselines for evaluation.
Instead of asking ChatGPT to identify the node associated to each subtask,
we ask for the subtasks and nodes given the query-able scene graph and the high-level task.
The results are shown in Table~\ref{tab:task_eval}.
\name outperforms the baselines by a large margin.
The GPT-powered baselines tend to provide a large number of subtasks that include
many contextually relevant but redundant and sometimes tangential subtasks, and this can be seen in particular with the lower subtask precision for Hydra with GPT.
HOV-SG is not well suited to parsing high-level tasks,
as it expects a formatted hierarchical query that is aligned with the scene graph structure.

\begin{table}%
    \small
    \setlength{\tabcolsep}{2pt}
    \centering
    \begin{tabular}{c ccc}
    \toprule
    Method & s-rec (\%) & s-prec (\%) & t-acc (\%) \\
    \midrule
    Hydra + GPT & 9.43 & 15.51 & 4.88 \\
    HOV-SG + GPT & 4.55 & 4.87 & 1.95 \\
    ASHiTA & \bf{10.39} & \bf{20.6}& \bf{9.27} \\
    \rowcolor{gray!20}
    ASHiTA (GT Pos) & 15.12 & 34.47 & 16.59 \\
    \rowcolor{blue!10}
    Hydra (GT Seg)~\cite{Hughes22rss-hydra} + GPT & 17.06 & 18.98 & 14.63 \\
    \rowcolor{blue!10}
    ASHiTA (GT Pos + Txt Emb) & \bf{38.71} & \bf{34.39} & \bf{36.1}\\
    \midrule 
    \end{tabular}
    \caption{Evaluation of Hierarchical Task Analysis.
    Gray highlight indicates directly using ground-truth object positions. Blue highlight marks trials having knowledge of ground-truth labels.\label{tab:task_eval}}\vspace{-6mm}
\end{table}



\subsection{System Ablations}
\label{sec:ablation}
We perform ablations on the components of \name,
comparing the full \name pipeline against (i) replacing the \hib with recursively running the original IB algorithm,
(ii) removing the Top-Down pruning step (Sec.~\ref{sec:scene_hierarchy}), (iii) removing the Spatial Update (Sec.~\ref{sec:task_update}),
and (iv) removing the LLM-based Hierarchy Refinement (Sec.~\ref{sec:task_update}).
We also include a comparison against using only GPT and the \primitives: in this case, we first obtain words in the word bank based on the max cosine similarity score,
and query \texttt{GPT-4o-mini} with the labeled primitives to come up with the subtasks and associated grounding.
The results are shown in Table~\ref{tab:system_ablations}.
While applying IB recursively compared to \hib yields a higher precision,
we obtain significantly lower recall due to the disconnect between layers of the task hierarchy in recursive IB variant.
\begin{table}%
    \small
    \setlength{\tabcolsep}{2pt}
    \centering
    \begin{tabular}{c cccc}
    \toprule
    Config & s-rec (\%) & s-prec (\%) & t-acc (\%) \\
    \midrule
    Recursive IB & 1.51 & \bf{24.53} & 1.46 \\
    w/o Top Down Pruning  & 9.22 & 18.93 & 5.37 \\
    w/o Spatial Update & 8.7 & 22.22 & 6.34 \\
    w/o Hierarchy Refinement & 7.71 & 23.13 & 6.83 \\
    Primitives + GPT & 6.14 & 7.16 & 5.37 \\
    ASHiTA & \bf{10.39} & 20.6 & \bf{9.27} \\
    \midrule 
    \end{tabular}
    \caption{Ablation study on components of the \name pipeline. \label{tab:system_ablations}}\vspace{-5mm}
\end{table}

\subsection{Qualitative Results}
\label{sec:qualitative}

\begin{figure}[t]
\centering
\includegraphics[width=1.0\columnwidth]{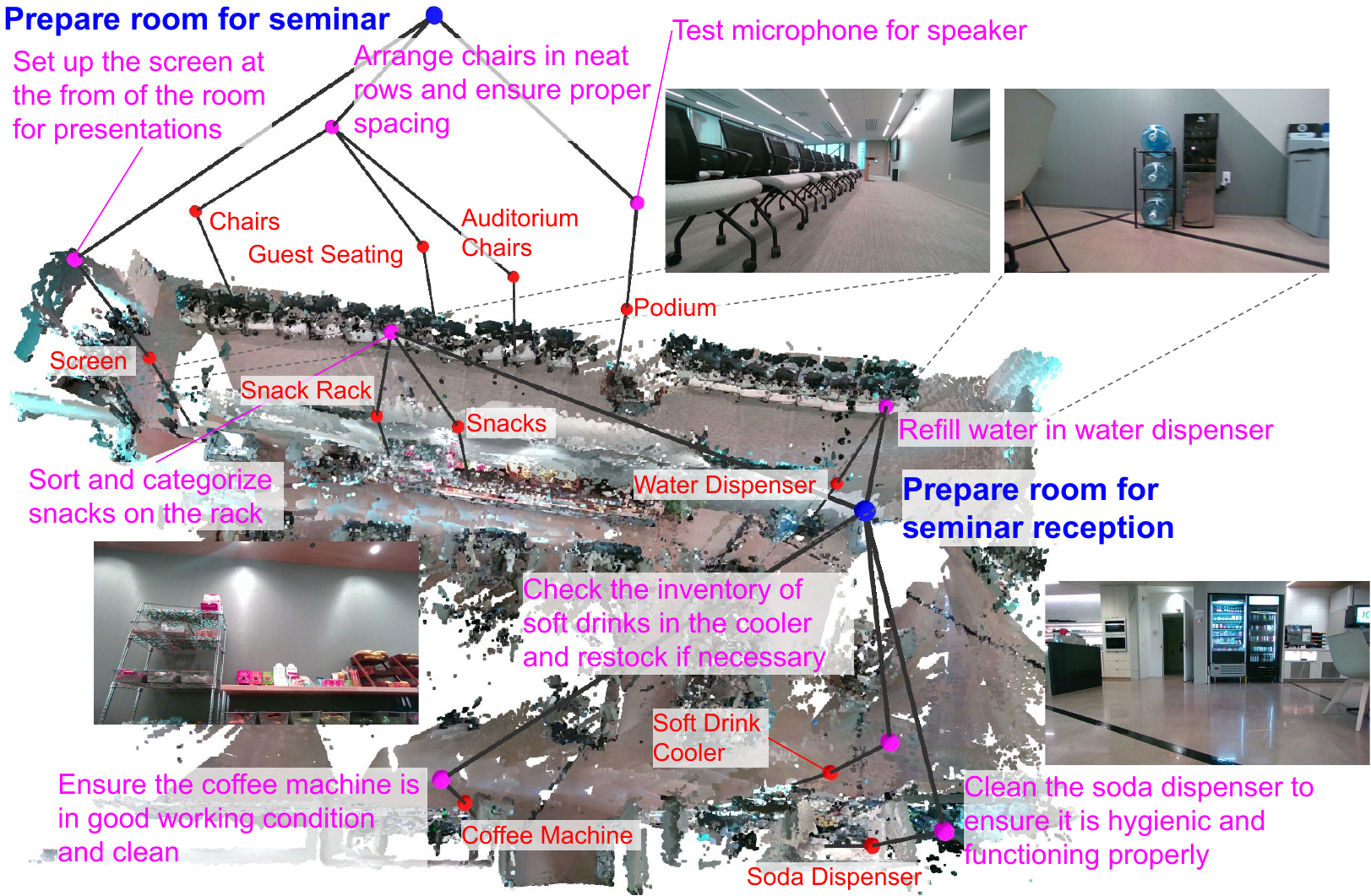}
\caption{\name demonstrated in a real-world seminar room and snack bar on a robot, given two high-level tasks.
Blue denotes the high-level tasks,
magenta the decomposed subtasks,
and red the items.}
\label{fig:snackbar_auditorium}
\vspace{-3mm}
\end{figure}

\begin{figure}[t]
\centering
\includegraphics[width=1.0\columnwidth]{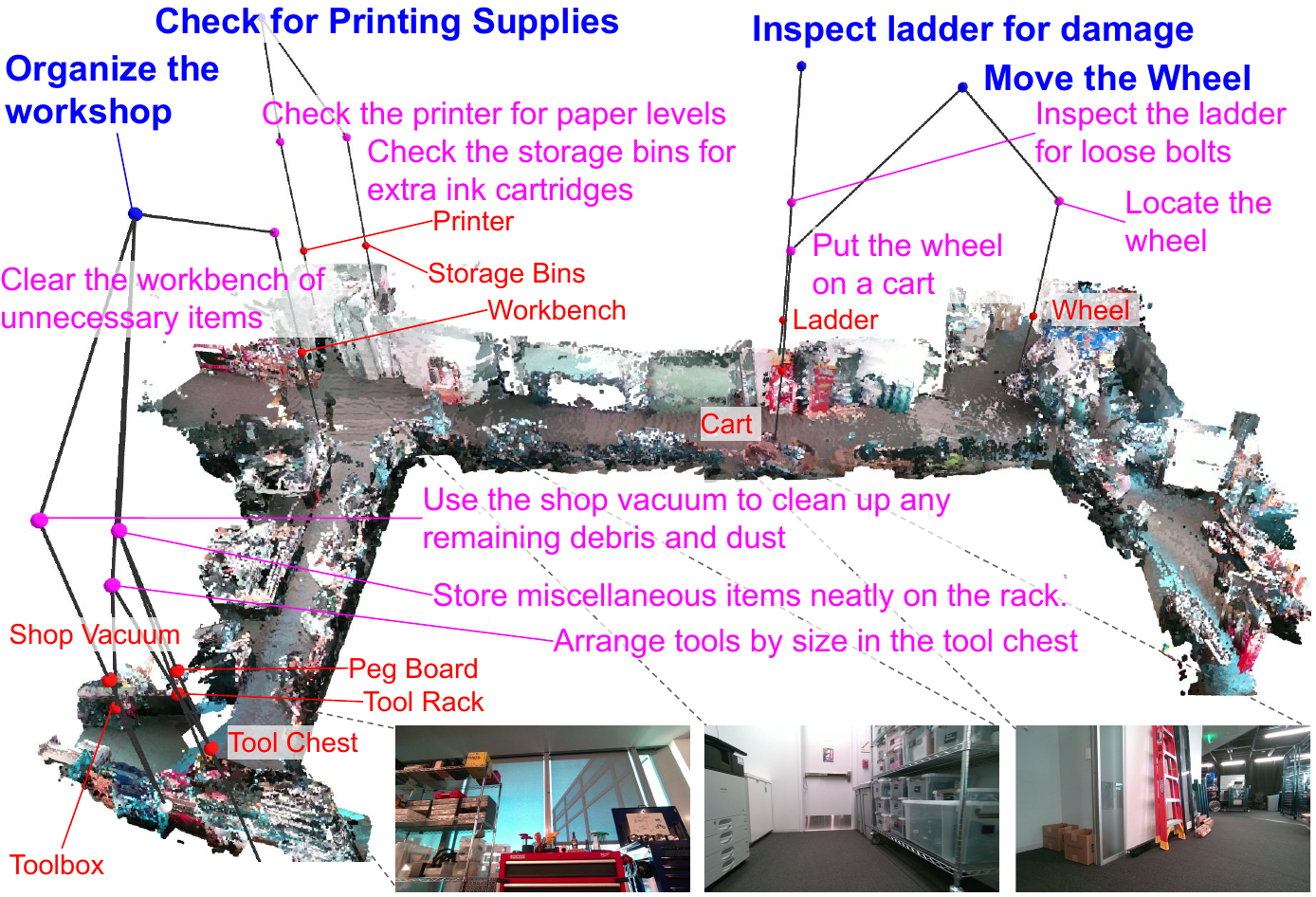}
\caption{\name demonstrated in a real-world hardware workshop environment on a robot
given 4 high-level tasks.}
\label{fig:hardware_shop}
\vspace{-5mm}
\end{figure}

\myParagraph{Real-World Demonstrations}
We demonstrate \name working in real-world environments.
In Fig.~\ref{fig:snackbar_auditorium}, a Boston Dynamics Spot is teleoperated through a snack bar and an adjacent seminar room.
Given high-level tasks of ``prepare room for a seminar'' and ``prepare room for seminar reception'',
\name was able to detect grounded items and subtasks associated with these high-level tasks.
One limitation of \name exposed here is the ability to deal with multiple identical items.
As seen under the chair arrangement subtask in Fig.~\ref{fig:snackbar_auditorium}, 
even though there are many identical chairs in the seminar room,
\name retained only three chairs and assigned them unique labels.
In Fig.~\ref{fig:hardware_shop}, the Spot explores a hardware area in an office building. \name is given four high level tasks of varying abstraction,
is able to perform reasonable hierarchical task analyses on the given tasks,
and demonstrate the ability to incorporate additional subtasks to account for observed scene elements.

\section{Limitations and Future Work}
\label{sec:limitations}
One major limitation of \name is the lack of relations in \name's 3D scene graph,
which means that \name is unable to accurately handle the incorporation of spatial specifications like ``inside-of'' or ``on-top-of''.
Additionally, the level of detail of the task hierarchy is limited by the descriptiveness of the word bank.
Similar but distinct objects (\eg, different types of cups) will not be differentiated in the task hierarchy and the final scene graph if the word bank consists only of a general category type (\eg, cup).
Lastly, there is no guarantee that the subtasks in the generated task hierarchy are sufficient to carry out the given high-level task. 
The hierarchical tree structure also constrains the granularity of subtasks in an object-centric manner,
preventing an object from being shared across multiple subtasks.
A possible future direction is to incorporate classical planning approaches to verify and refine the generated subtasks~\cite{Chen23icra-AutoTAMP}.

\section{Conclusion}
\label{sec:conclusion}
We present \name, the first framework to generate a task hierarchy grounded to a 3D scene graph
by breaking down high-level tasks into grounded subtasks.
To achieve this, we introduce the Hierarchical Information Bottleneck (\hib),
a hierarchical generalization of the well known Information Bottleneck algorithm~\cite{Tishby01accc-IB},
and present an iterative approach that alternates between hierarchical task analysis and scene graph generation.
We demonstrated that \name is able to decompose high-level tasks into grounded subtasks
better than LLM and scene graph baselines by benchmarking against human-annotated scene-specific task breakdowns.

\section*{Acknowledgments}

The authors gratefully acknowledge Dominic Maggio for insightful discussions on VLMs and task-driven representations.
We also thank Xiaohan Zhang, Ashay Athalye, Yixuan Wang, Jianing Qian, and many others at the Robotics and AI Institute for their thoughtful conversations on language and planning for robotics
that contributed to the formulation of this work.
{
    \small
    \bibliographystyle{ieeenat_fullname}
    \bibliography{refs, myRefs, staged_refs}
}


\clearpage
\setcounter{page}{1}
\maketitlesupplementary

\begin{figure}[ht]
\centering
\includegraphics[width=1.0\columnwidth]{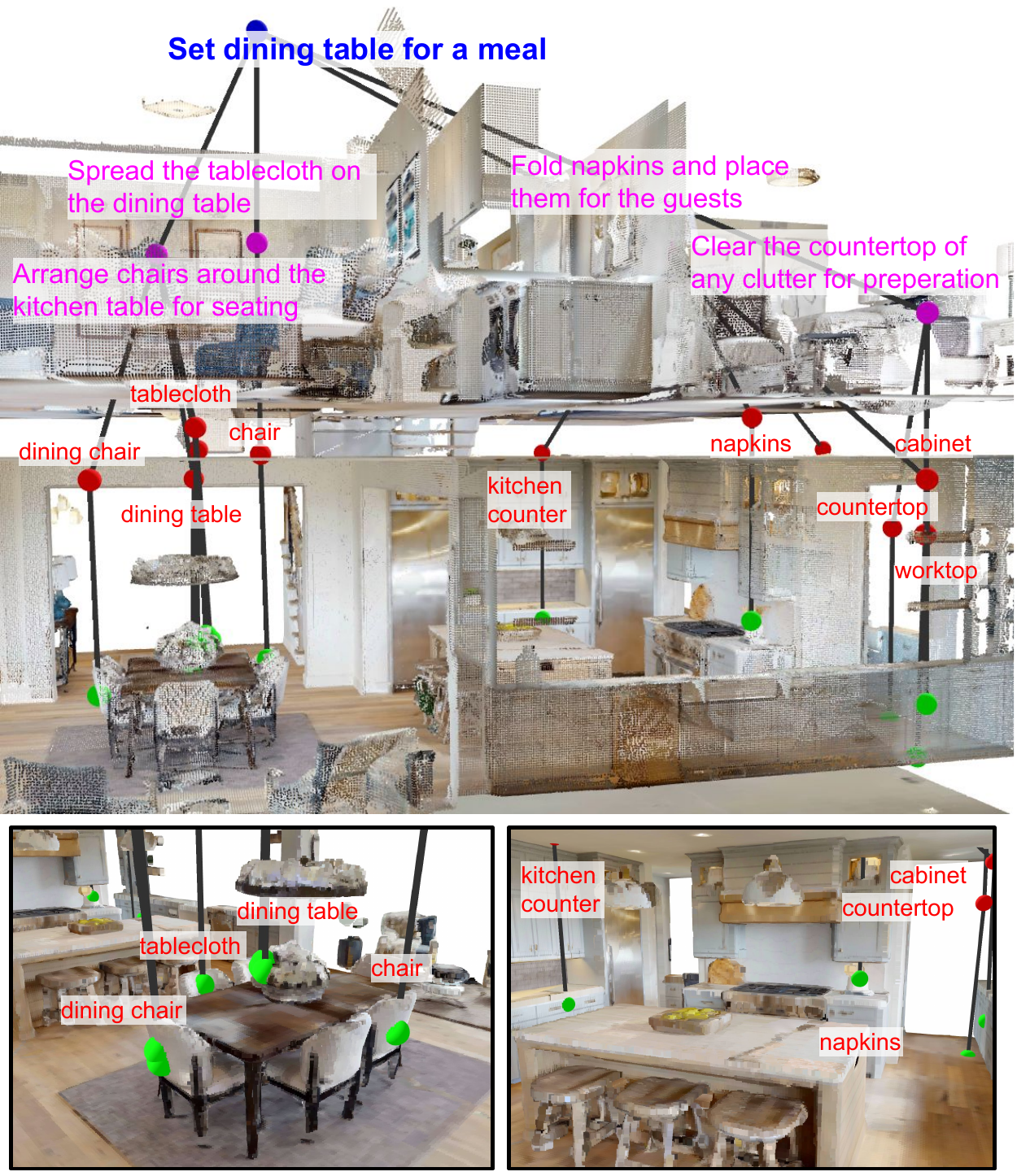}
\caption{\name given the high-level task of "set dining table for a meal" from HM3DSem scene 00862-LT9Jq6dN3Ea~\cite{Yadav22-habitatchallenge}.
Green markers are the primitives, red the items, magenta the subtasks, and blue the given high-level task.
The bottom two frames show the zoomed in views of the scene graph and scene.}
\label{fig:set_dining_table}
\end{figure}

\begin{figure}[ht]
\centering
\includegraphics[width=0.9\columnwidth]{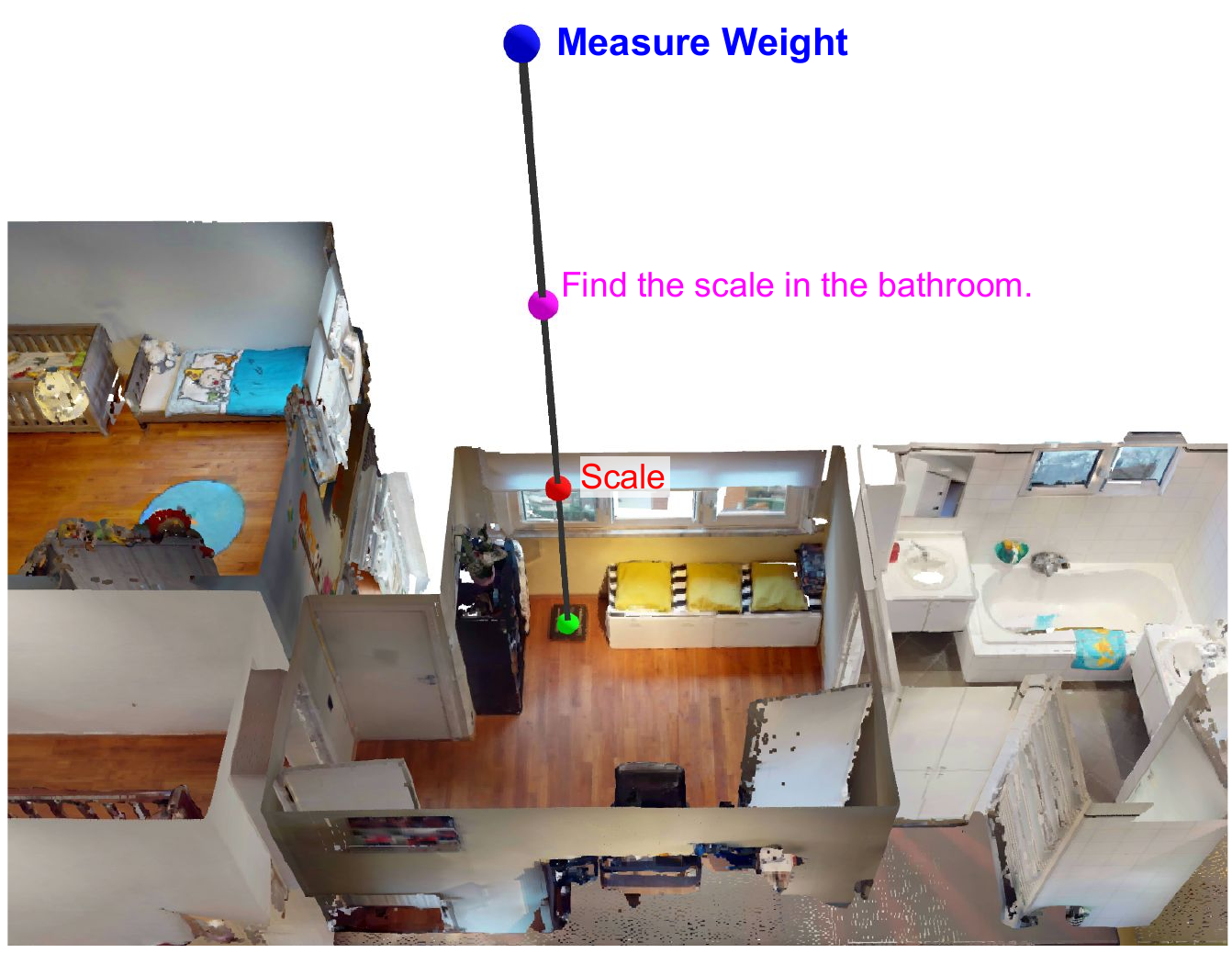}
\caption{\name given the high-level task of "measure weight" from HM3D scene 00890-6s7QHgap2fW~\cite{Yadav22-habitatchallenge}.
Green marks the primitive, red the item, magenta the subtask, and blue the given high-level task.}
\label{fig:measure_weight}
\end{figure}

\begin{figure}[ht]
\centering
\includegraphics[width=1.0\columnwidth]{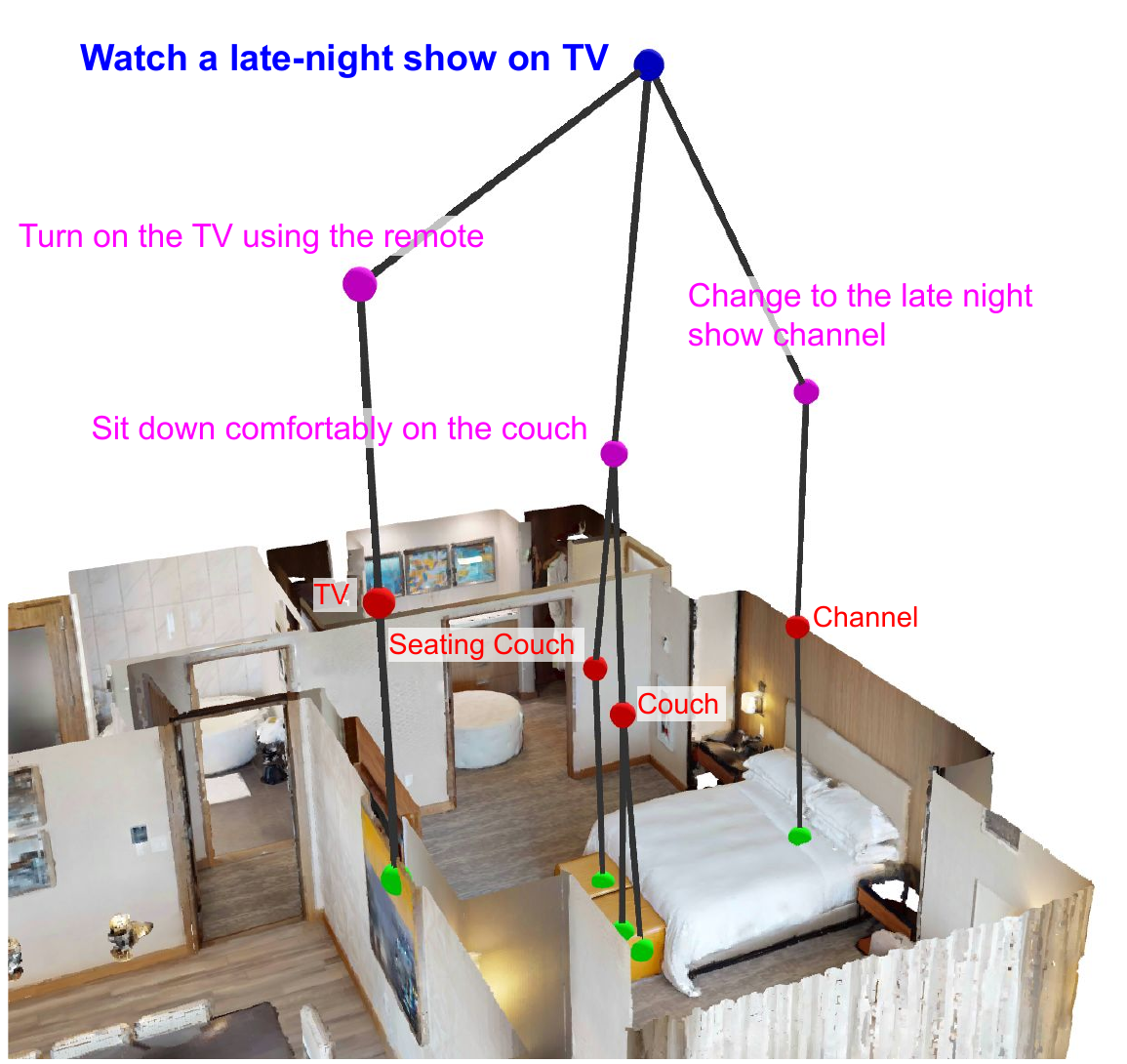}
\caption{\name given the high-level task of "Watch a late-night show on TV" from HM3DSem scene 00829-QaLdnwvtxbs~\cite{Yadav22-habitatchallenge}.
Green markers are the primitives, red the items, magenta the subtasks, and blue the given high-level task.}
\label{fig:watch_latenight_tv}
\end{figure}

\begin{figure}[htpb!]
\centering
\includegraphics[trim=0 0 0 0, clip, width=0.9\columnwidth]{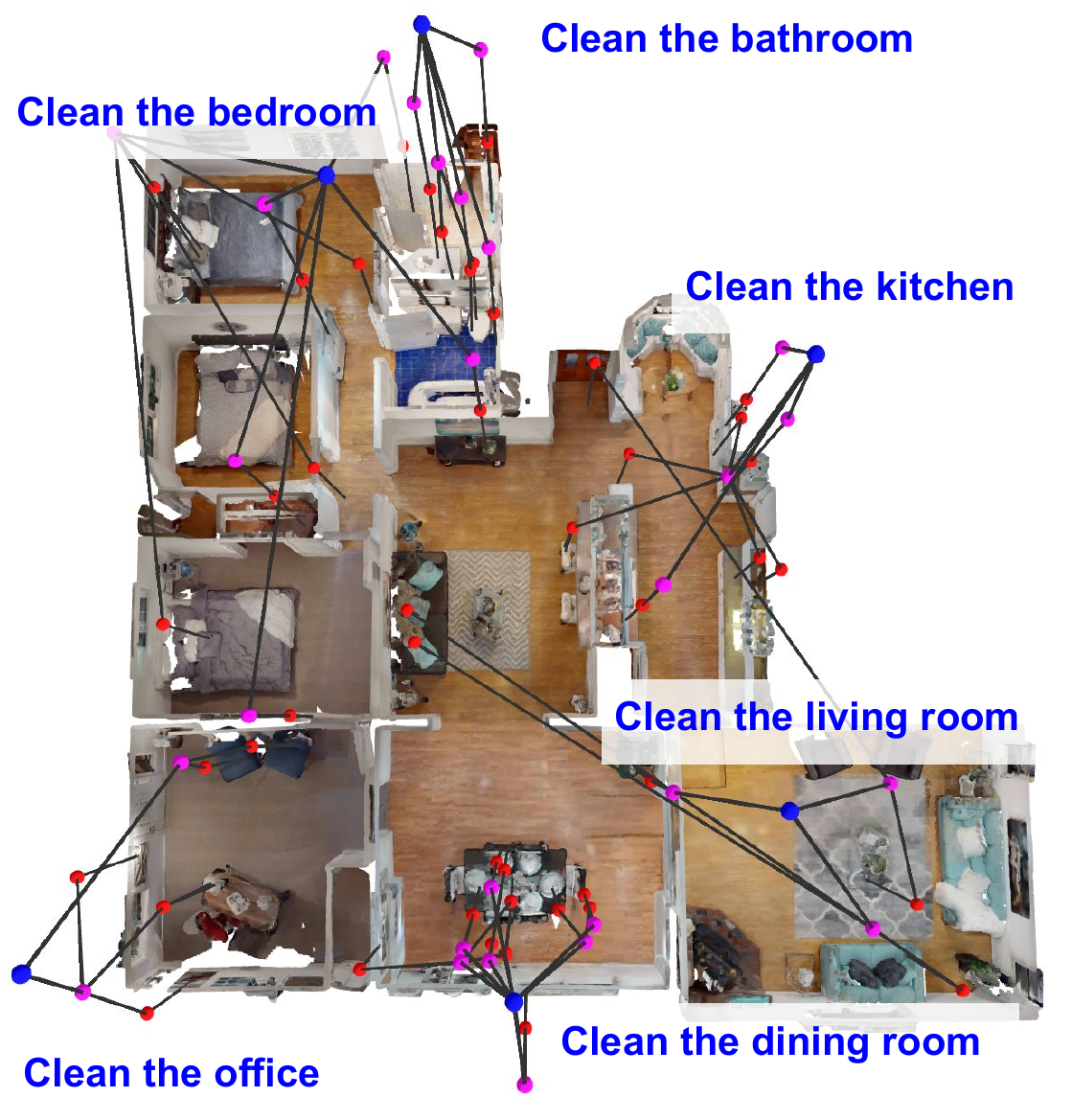}
\caption{Given a set of high-level tasks that directly relate to the rooms in a house,
we can approximately recover similar entities to that of prior scene graph construction approaches~\cite{Hughes22rss-hydra, Werby24rss-hovsg}.
For visualization clarity, we only show the labels of the high-level tasks. The purple nodes mark the subtasks and the red the associated objects.}
\label{fig:rooms_and_objects}
\end{figure}

\begin{figure*}[t]
    \centering
    \begin{subfigure}[b]{0.99\textwidth}
        \centering
        \includegraphics[width=\linewidth]{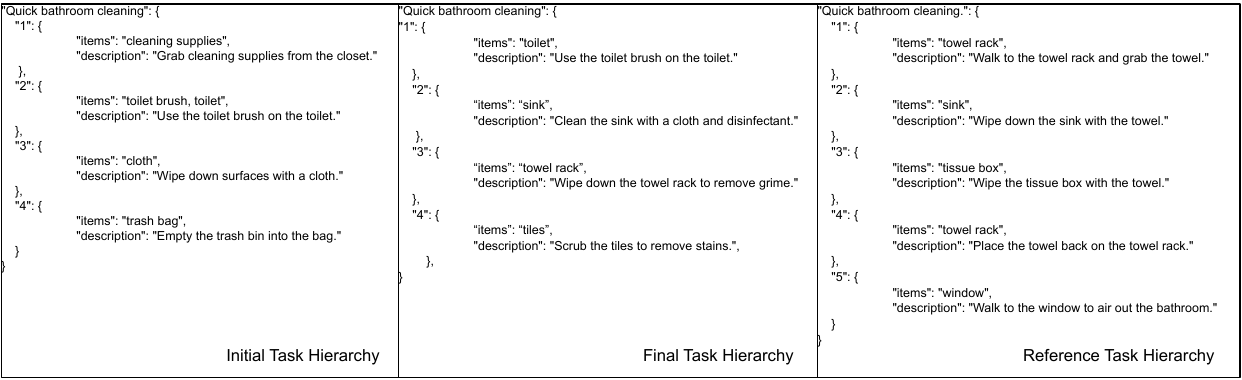} 
        \caption{Initial task hierarchy, final \name generated task hierarchy, and the reference task hierarchy from \cite{zhang24arxiv-sg3d}.}
        \label{fig:detailed_task_hierarchy}
    \end{subfigure}

    \vskip\baselineskip 

    \begin{subfigure}[b]{0.49\textwidth}
        \centering
        \includegraphics[width=\linewidth]{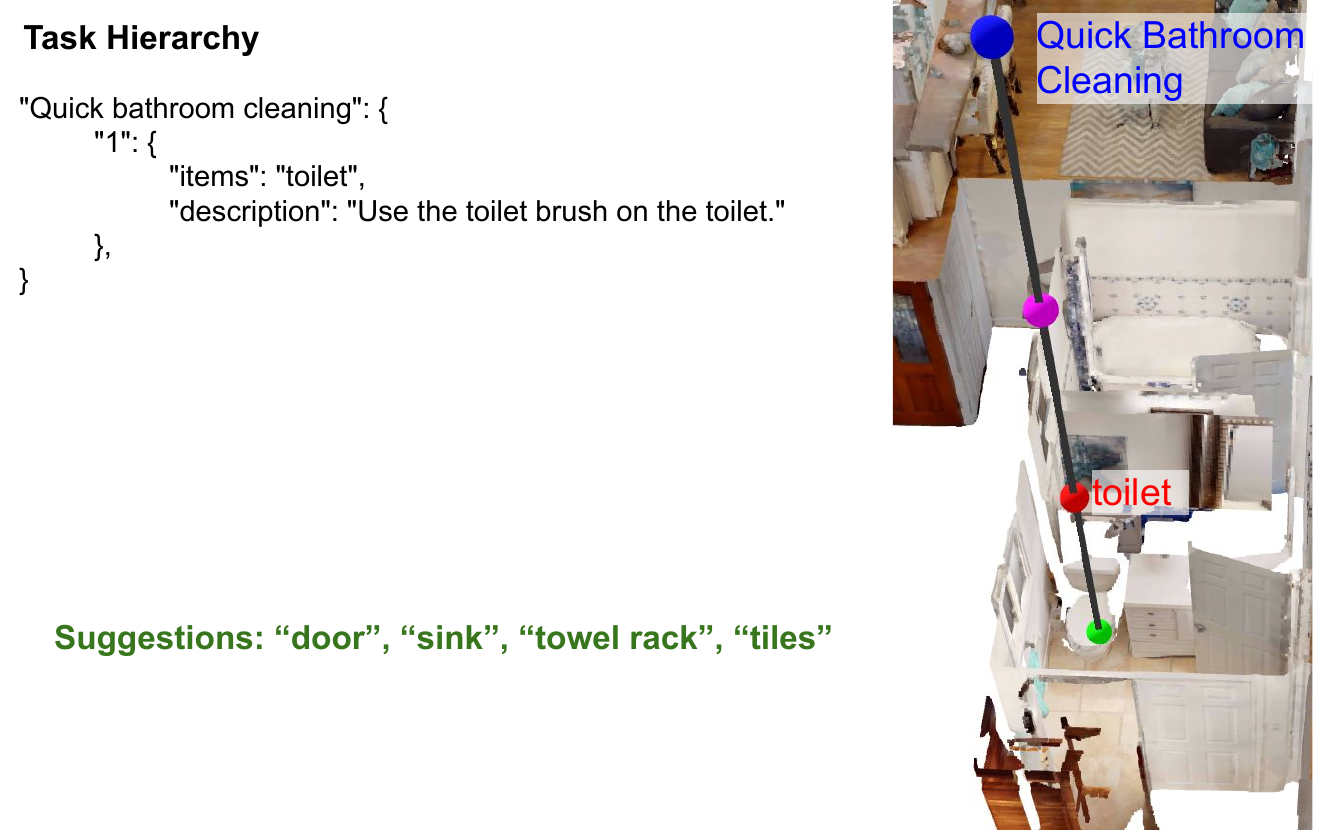} 
        \caption{Task hierarchy and scene graph after first iteration.}
        \label{fig:first_iteration}
    \end{subfigure}
    \hfill
    \begin{subfigure}[b]{0.49\textwidth}
        \centering
        \includegraphics[width=\linewidth]{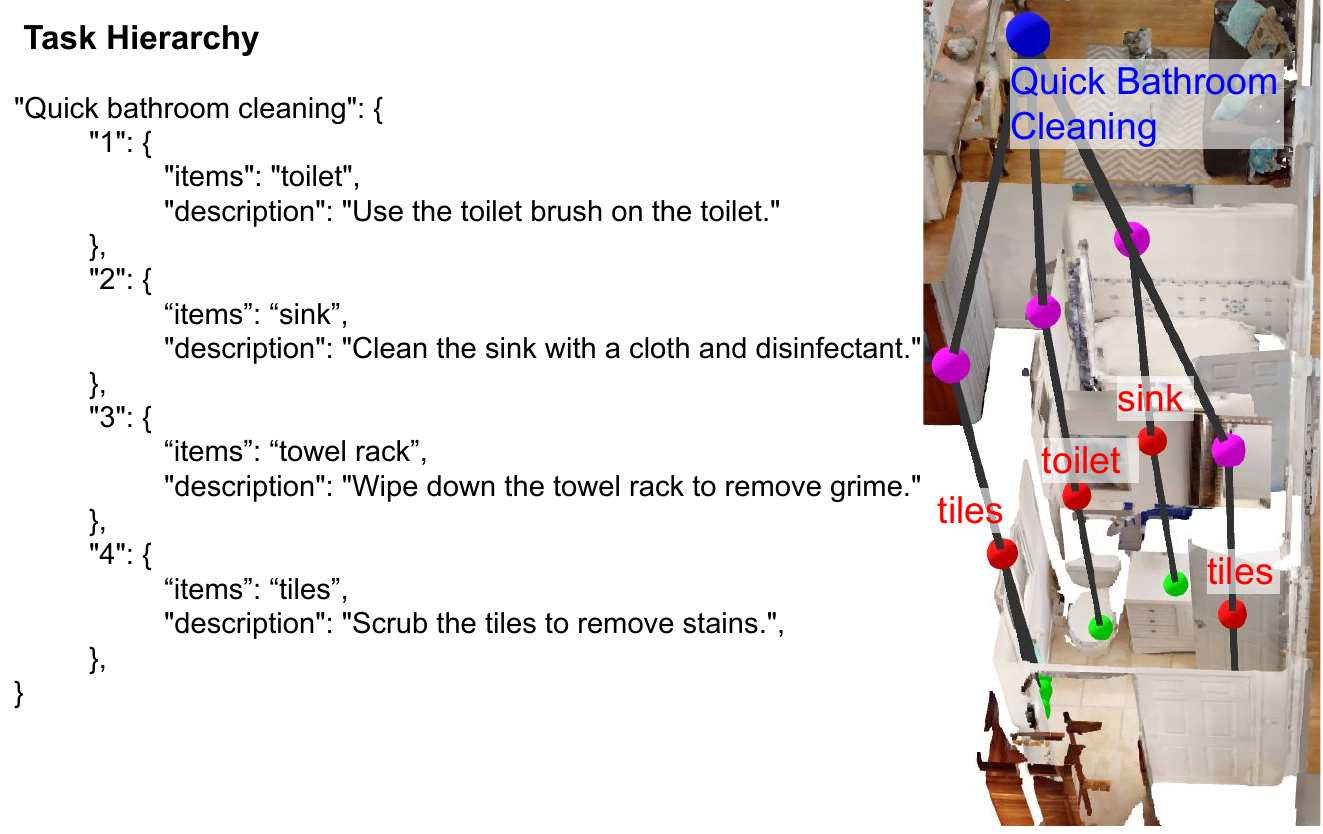}
        \caption{Task hierarchy and scene graph after second iteration.}
        \label{fig:second_iteration}
    \end{subfigure}
    
    \caption{\name with two iterations for the high-level task of "Quick bathroom cleaning". (b) From the initial hierarchy, only the toilet is successfully grounded in the generated scene graph. From this iteration, the suggestions of "door", "sink", "towel rack", and "tiles" are generated.
    These suggestions are used to update the task hierarchy, successfully recovering part of the human-annotated reference task hierarchy
    (recall that our evaluation is object-centric, as discussed in Sec.~\ref{sec:experiment}).}
    \label{fig:ashita_bathroom_cleaning}
\end{figure*}

\begin{figure*}[htpb!]
    \centering
    \begin{subfigure}[b]{0.99\textwidth}
        \centering
        \includegraphics[width=\linewidth]{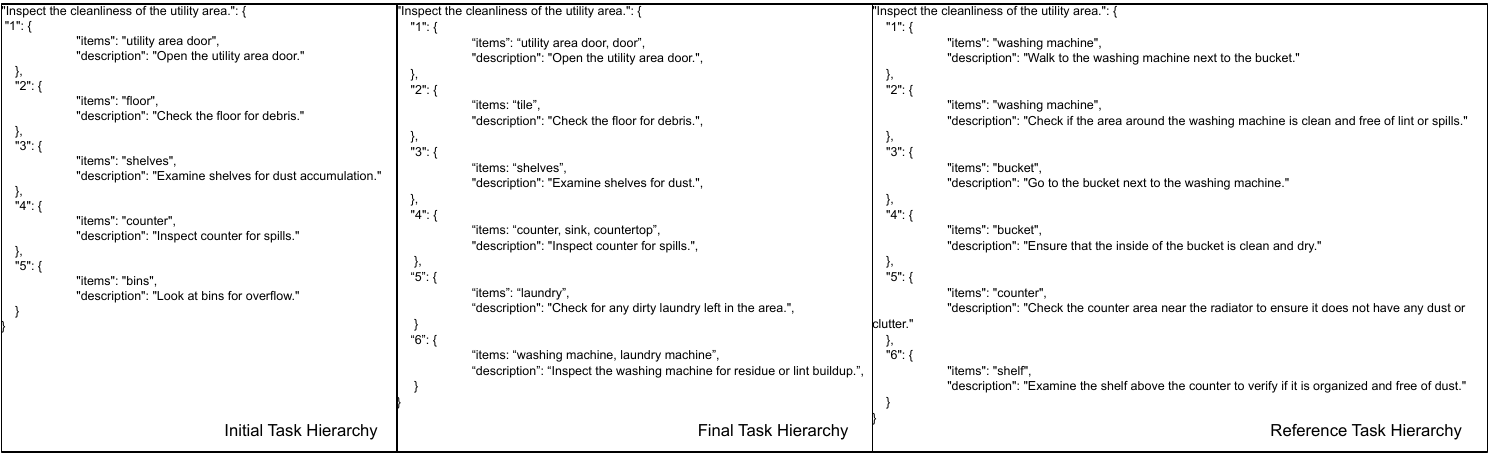} 
        \caption{Initial task hierarchy, final \name generated task hierarchy, and the reference task hierarchy from \cite{zhang24arxiv-sg3d}.}
        \label{fig:detailed_task_hierarchy}
    \end{subfigure}

    \vskip\baselineskip 

    \begin{subfigure}[b]{0.49\textwidth}
        \centering
        \includegraphics[width=\linewidth]{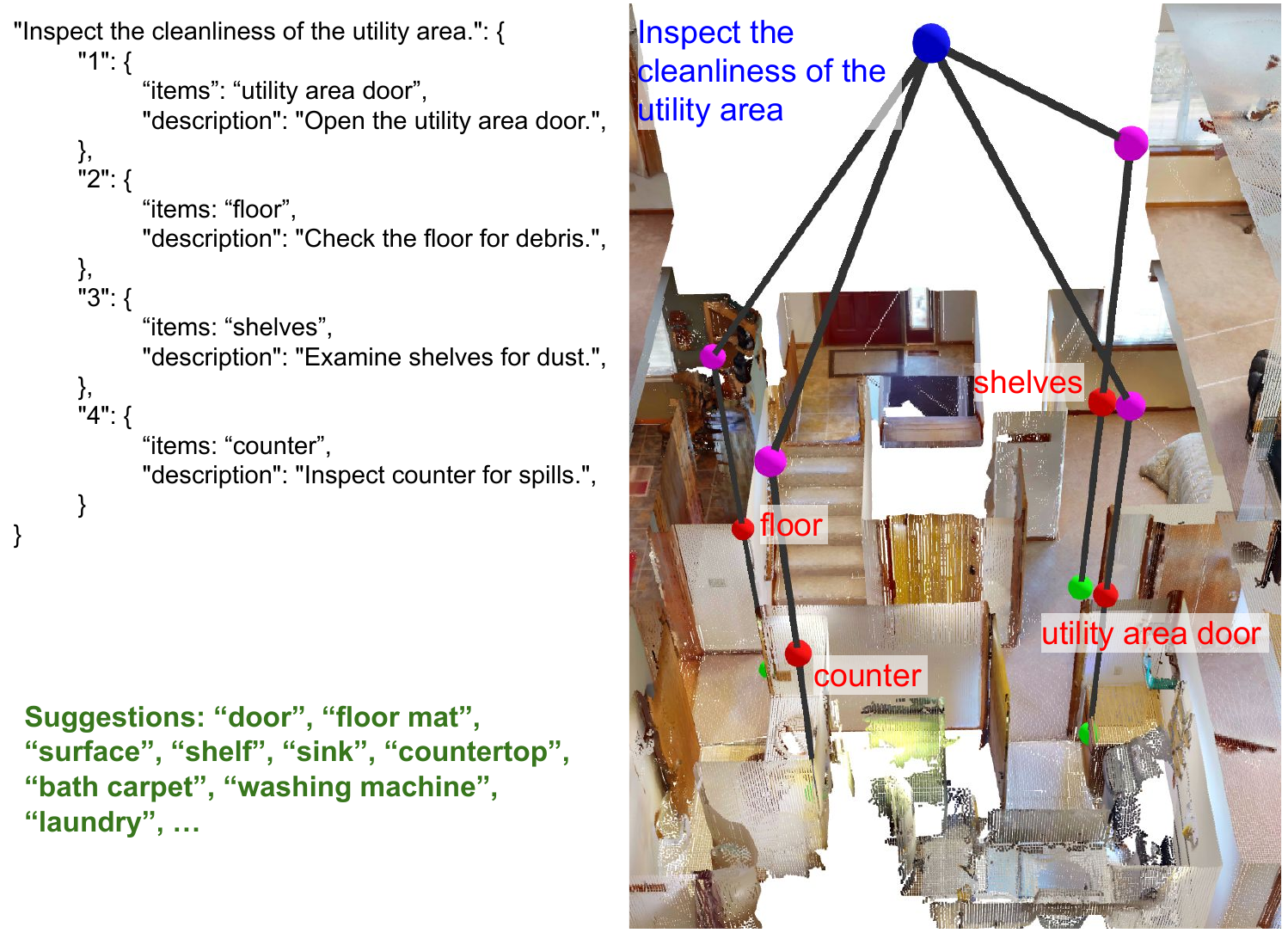} 
        \caption{Task hierarchy and scene graph after first iteration.}
        \label{fig:util_first_iteration}
    \end{subfigure}
    \hfill
    \begin{subfigure}[b]{0.49\textwidth}
        \centering
        \includegraphics[width=\linewidth]{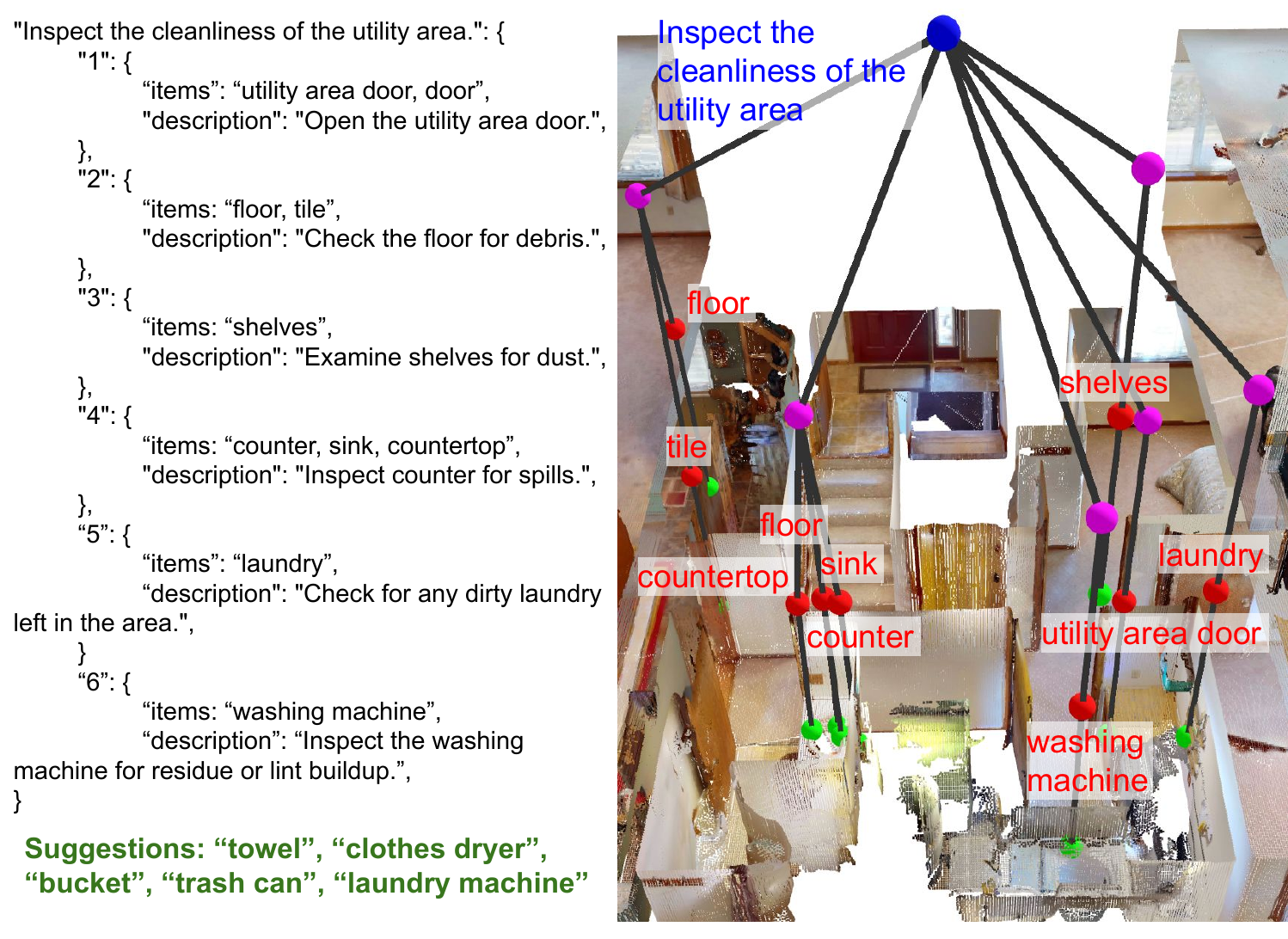}
        \caption{Task hierarchy and scene graph after second iteration.}
        \label{fig:util_second_iteration}
    \end{subfigure}
    \vskip\baselineskip 
    \begin{subfigure}[b]{0.49\textwidth}
        \centering
        \includegraphics[width=\linewidth]{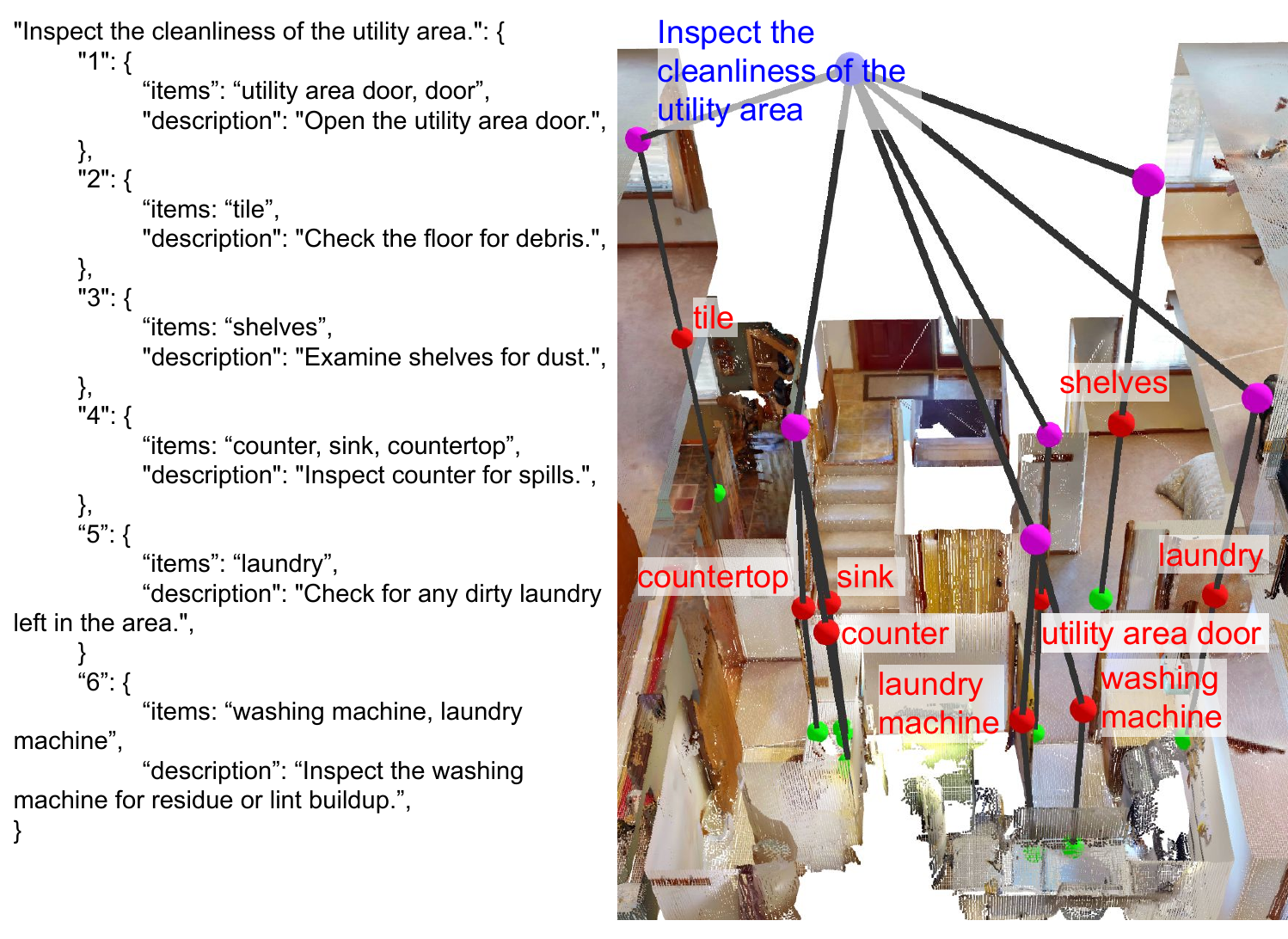}
        \caption{Task hierarchy and scene graph after third iteration.}
        \label{fig:util_third_iteration}
    \end{subfigure}
    \hfill
    \begin{subfigure}[b]{0.49\textwidth}
        \centering
        \includegraphics[width=\linewidth]{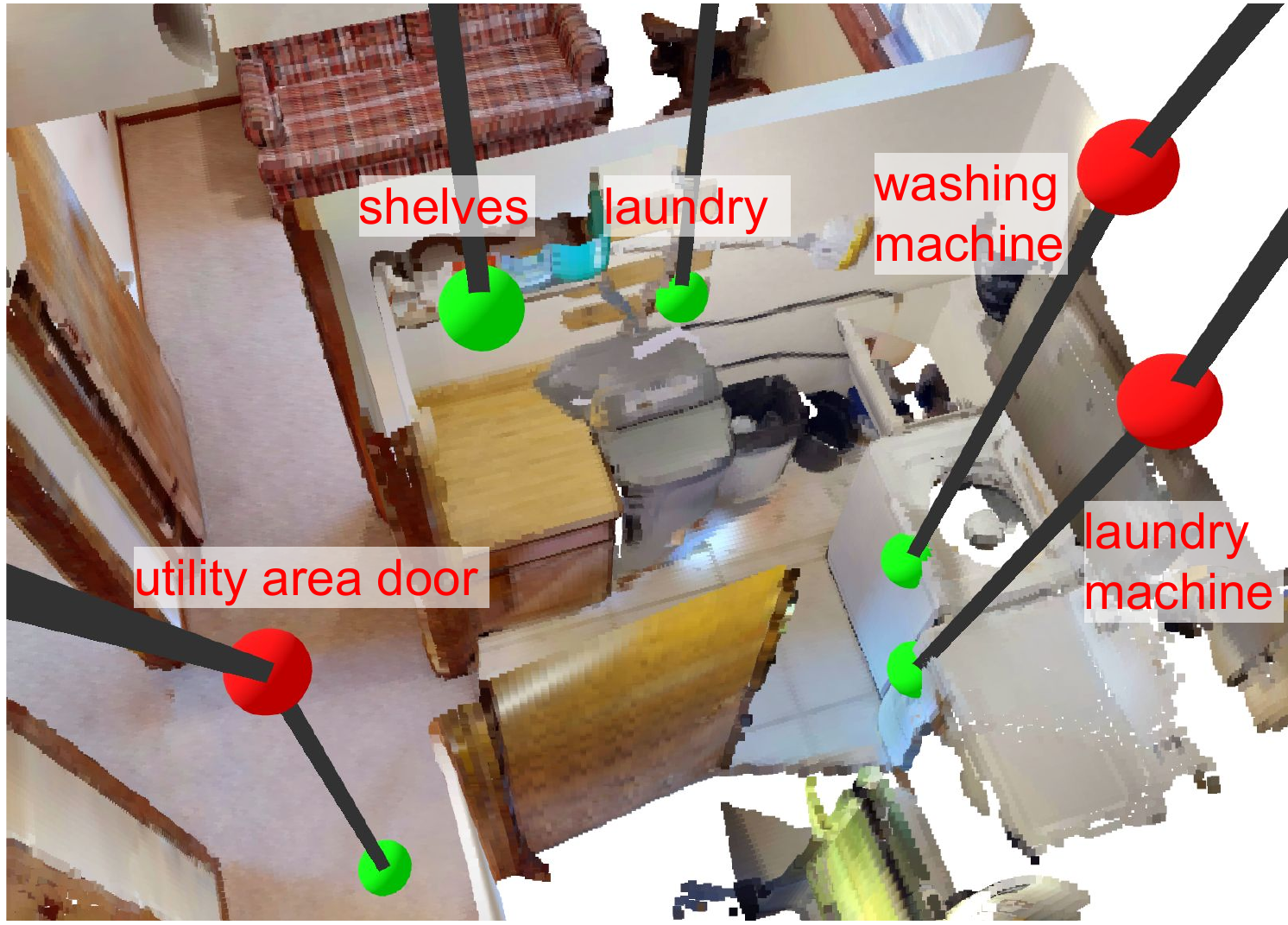}
        \caption{Zoomed in view on the primitives and the scene.}
        \label{fig:util_zoom}
    \end{subfigure}
    
    \caption{\name with three iterations for the high-level task of "Inspect the cleanliness of the utility area". (b) From the initial hierarchy, all items except for the bins are grounded in the generated scene graph. Many suggestions are generated to update the hierarchy.
    (c) Suggested items "laundry" and "washing machine" are used to generate new subtasks, these are also successfully grounded in the generated scene graph. Grounding these items trigger a few additional suggestions.
    (d) Suggested items "laundry machine" added to the task hierarchy. This refers to the dryer next to the washing machine. Our estimated final task hierarchy recalls all of the grounded items in the reference task hierarchy except for the bucket next to the washing machine.
    \label{fig:ashita_utility_inspect}}
\end{figure*}

\section{Qualitative Examples}
\label{supp:extra_results}

\subsection{SG3D Hierarchical Task Analysis}

We include some qualitative examples of \name with the SG3D~\cite{zhang24arxiv-sg3d} high-level tasks in the HM3DSem~\cite{Yadav22-habitatchallenge} dataset.
In Fig.~\ref{fig:measure_weight}, \name is given a single high-level task of "measure weight", and is able to correctly identify the scale.
Note that \name is an object-centric approach and does not refine or evaluate on the specific \emph{description} of the subtasks.
In Fig.~\ref{fig:set_dining_table} we show \name at a larger scale in a kitchen and a dining room for "set dining table for a meal",
\name comes up with reasonable subtasks and identified relevant items associated with the task.
Lastly, in Fig.~\ref{fig:watch_latenight_tv}, the task given is "watch late-night show on TV",
\name also mostly identifies the correct subtasks and items.
A few mistakes are: selecting the TV in the neighboring dining room instead of the bedroom,
and associating the non-item word "channel" to the bed.
Note that \name is also able to cluster the two primitives (green) together for the "couch" item.

\subsection{Detailed Example of \name}
In this section, we demonstrate in detail the alternating iterations of \name with specific examples from the SG3D dataset.
In Fig.~\ref{fig:ashita_bathroom_cleaning} we show the comparison between the initial task hierarchy given to \name, the final task hierarchy, 
and the reference task hierarchy from SG3D~\cite{zhang24arxiv-sg3d} alongside two iterations of \name.
In the first iteration, only the subtask related to the toilet is grounded in the scene graph,
but incorporating the suggestions from this iteration, the task hierarchy is refined and we end up with four grounded subtasks.
In Fig.~\ref{fig:ashita_utility_inspect} we give \name the task of "inspect the cleanliness of the utility area".
With three iterations, \name is able to incorporate a more complete grounded set of relevant objects and subtasks.

\subsection{Rooms and Objects}
While \name is not designed to recover traditional geometric structures such as room layouts
that other prior scene graph construction approaches do~\cite{Hughes24ijrr-hydraFoundations,Gu24icra-conceptgraphs,Werby24rss-hovsg},
with a specially chosen set of high-level tasks, we can approximately recover a similar set of rooms and objects.
This is shown in Fig~\ref{fig:rooms_and_objects},
where \name is given 6 high-level cleaning tasks related to the types of rooms in the scene.
However, it is clear that
\name lacks the understanding of structures and the overall floor plan,
and only retains the entities that are deemed relevant to the given tasks.

\section{Initial Hierarchy Ablations}
\label{supp:initial_hierarchy_ablation}
In Section~\ref{sec:experiment},
we use \texttt{GPT-4o-mini} to generate the initial task hierarchy by first giving \texttt{GPT-4o-mini}a manually generated task hierarchy for some arbitrary task as an example then querying with the prompt:
\begin{framed}
\begin{quote}
    "Given the example above, generate a concise task hierarchy for <task>, ensuring brief and clear descriptions."
\end{quote}
\end{framed}
            
Here we include an ablation to evaluate the impact of the initial task hierarchy.
Using three different ChatGPT generated initial hierarchies
including one privileged with the inclusion of objects in the scene as priors.
The results are shown in Table ~\ref{tab:initialization}.
The impact of the initial task hierarchy is minimal, even with privileged priors,
and does not change the reported trend of the results shown in Table~\ref{tab:task_eval}.

\begin{table}%
    \scriptsize
    \setlength{\tabcolsep}{2pt}
    \centering
    \begin{tabular}{c c ccc}
    \toprule
    Initial Hierarchy & Method & s-rec (\%) & s-prec (\%) & t-acc (\%) \\
    \midrule
    \multirow{2}{*}{1} & ASHiTA & 10.39 & 20.6 & 9.27 \\
                        & ASHiTA (GT Pos + Txt Emb) & 38.71 & 34.39 & 36.1 \\
    \midrule
    \multirow{2}{*}{2} & ASHiTA & 9.95 & 20.41 & 7.8 \\
                       & ASHiTA (GT Pos + Txt Emb) & 40.56 & 35.38 & 37.56 \\
    \midrule
    \rowcolor{gray!20}
     & ASHiTA & 14.3 & 17.0 & 12.2 \\
    \rowcolor{gray!20} \multirow{-2}{*}{Privileged} & ASHiTA (GT Pos + Txt Emb) & 42.13 & 38.68 & 42.93 \\
    \midrule
    \end{tabular}
    \caption{\name and \name with ground-truth objects and labels with different initial task hierarchies.
    Top two rows are generated with ChatGPT given only the abstract tasks; the last row was generated given ground-truth labels of known objects in the scene.\label{tab:initialization}}\vspace{-3mm}
\end{table}

\section{Derivation of the Hierarchical Information Bottleneck}
\label{supp:hib}

In this section, we detail the derivation of the iterative multi-layer update steps~\eqref{eq:hib_cases}
used to minimize the \hib functional in~\eqref{eq:hib},
following the same approach as the original IB derivation
outlined in \cite{Tishby01accc-IB}.
To do this, we first formulate the Lagrangian for \hib along with the derivative of the Lagrangian.
Next, we express the derivatives of the conditional probabilities used in \hib
conditioned on the Markov chain assumption.
These expressions can be substituted into the derivative of the Lagrangian,
which allows us to solve for the zero of the Lagrangian derivative.

Accounting for the constraint that $\prob{S_k|S_k-1}$ is a valid probability,
the Lagrangian of \eqref{eq:hib} can be written as,
\begin{equation}\label{eq:lagrangian}
\begin{split}
    \mathcal{L} &= \sum_{i=1}^n \{I(\sgraph_{i-1} ; \sgraph_i) - \beta I(\tasks_i ; \sgraph_i) \\
    &+ \sum_{s_{i-1} \in \sgraph_{i-1}}\lambda (s_{i-1})[\sum_{s_i \in \sgraph_{i}}p(s_i|s_{i-1}) - 1]\}\\
\end{split}
\end{equation}

Recalling the definition of mutual information
\begin{equation}\label{eq:mutual_information}
\begin{split}
    I(X; Y) &= \sum_{x}\sum_{y}p(x, y) log(\frac{p(x, y)}{p(x) p(y)})\\
    &= \sum_{x}\sum_{y}p(x|y)p(y) log(\frac{p(x|y)}{p(x)})\\
\end{split}
\end{equation}

using the logarithm properties, we expand \eqref{eq:lagrangian} as,
\begin{equation}\label{eq:lagrangian_expanded}
\begin{split}
    \mathcal{L} &= \sum_{i=1}^n \{ \\
    &\sum_{s_{i-1}} p(s_{i-1}) \sum_{s_i} p(s_i | s_{i-1}) [log(p(s_i|s_{i-1})) - log(p(s_i))] \\
    &-\beta\sum_{t_i} p(t_i) \sum_{s_i} p(s_i | t_i) [log(p(s_i|t_i)) - log(p(s_i))] \\
    &+\sum_{s_{i-1}}\lambda (s_{i-1})[\sum_{s_i}p(s_i|s_{i-1}) - 1)]\} \\
\end{split}
\end{equation}

Our goal is to derive $p(s_k|s_{k-1})$ for some arbitrary level $k$ for some integer $k \in [1, n]$.
We rewrite and expand \eqref{eq:lagrangian_expanded} and break up the sum of the levels into three parts:
from level $1$ to $k-1$, level $k$, and from level $k+1$ to level $n$.
The Markov chain assumption designate that the levels lower than $k$ are not dependent on $p(s_k|s_{k-1})$,
level $k$ is directly dependent on $p(s_k|s_{k-1})$,
and the higher levels are indirectly dependent on $p(s_k|s_{k-1})$.
This means that when we take the derivative of the Lagrangian with respect to $p(s_k|s_{k-1})$ for fixed $s_k$ and $s_{k-1}$,
the terms related to the first part are zero.
This broken up expression is as follows,
arranged in order of the three terms in \eqref{eq:lagrangian},
and for each term, broken up into three parts based on $k$ as described,

\begin{equation}\label{eq:lagrangian_expanded_k}
\begin{aligned}
    \mathcal{L} &= \sum_{i=1}^{k-1} \{ \\
    &\sum_{s_{i-1}} p(s_{i-1}) \sum_{s_i} p(s_i | s_{i-1}) [log(p(s_i|s_{i-1})) - log(p(s_i))] \} \\
    +&\sum_{s_{k-1}} p(s_{k-1}) \sum_{s_k} p(s_k | s_{k-1}) [log(p(s_k|s_{k-1})) - log(p(s_k))] \\
    +&\sum_{s_{k}} p(s_{k}) \sum_{s_{k+1}} p(s_{k+1} | s_{k}) [log(p(s_{k+1}|s_{k})) - log(p(s_{k+1}))] \\
    +&\sum_{i=k+2}^n \{ \\
    &\sum_{s_{i-1}} p(s_{i-1}) \sum_{s_i} p(s_i | s_{i-1}) [log(p(s_i|s_{i-1})) - log(p(s_i))] \} \\
    -&\beta\sum_{i=1}^{k-1} \{\sum_{t_i} p(t_i) \sum_{s_i} p(s_i | t_i) [log(p(s_i|t_i)) - log(p(s_i))] \} \\
    -&\beta\sum_{t_k} p(t_k) \sum_{s_k} p(s_k | t_k) [log(p(s_k|t_k)) - log(p(s_k))] \\
    -&\beta\sum_{i=k + 1}^n \{\sum_{t_i} p(t_i) \sum_{s_i} p(s_i | t_i) [log(p(s_i|t_i)) - log(p(s_i))] \} \\
    +&\sum_{i=1}^{k-1} \{\sum_{s_{i-1}}\lambda (s_{i-1})[\sum_{s_i}p(s_i|s_{i-1}) - 1)]\} \\
    +&\sum_{s_{k-1}}\lambda (s_{k-1})[\sum_{s_k}p(s_k|s_{k-1}) - 1)]\} \\
    +&\sum_{i=k+1}^n \{\sum_{s_{i-1}}\lambda (s_{i-1})[\sum_{s_i}p(s_i|s_{i-1}) - 1)]\} \\
\end{aligned}
\end{equation}

We can then take the derivative of the Lagrangian $\frac{\delta \mathcal{L}}{\delta p(s_k|s_{k-1})}$ 
for fixed $s_k$ and $s_{k-1}$.
From basic calculus, we can derive that
\begin{equation}\label{eq:basic_calculus}
    \frac{d(f(x) log(f(x)))}{dx} = \frac{df(x)}{dx}(log(f(x)) + 1)
\end{equation}

The terms related to the first part (level $1$ to $k-1$) are zero as explained above.
Using the chain rule along with \eqref{eq:basic_calculus}, the derivative of the Lagrangian~\eqref{eq:lagrangian_expanded_k} is,

\begin{equation}\label{eq:deriv_lagrangian}
\begin{split}
    &\frac{\delta \mathcal{L}}{\delta p(s_k|s_{k-1})}\\
    =&p(s_{k-1})[log(p(s_k|s_{k-1})) + 1] \\
    &- \frac{\delta p(s_k)}{\delta p(s_k|s_{k-1})}[log(p(s_k)) + 1] \\
    +& \frac{\delta p(s_k)}{\delta p(s_k|s_{k-1})}\sum_{s_{k+1}}p(s_{k+1} | s_k) log(p(s_{k+1}|s_k)) \\
    &- \sum_{s_{k+1}} \frac{\delta p(s_{k+1})}{\delta p(s_k|s_{k-1})}[log(p(s_{k+1})) + 1] \\
    +&\sum_{i=k+2}^n \{
    \sum_{s_{i-1}} \frac{\delta p(s_{i-1})}{\delta p(s_k|s_{k-1})} \sum_{s_i} p(s_i | s_{i-1}) log(p(s_i|s_{i-1})) \\
    &- \sum_{s_i}\frac{\delta p(s_i)}{\delta p(s_k|s_{k-1})}[log(p(s_i)) + 1] \} \\
    -&\beta\{ \sum_{t_k}p(t_k)\frac{\delta p(s_k|t_k)}{\delta p(s_k|s_{k-1})}[log(p(s_k|t_k)) + 1] \\
    &-\frac{\delta p(s_k)}{\delta p(s_k|s_{k-1})}[log(p(s_k)) + 1]) \}\\
    -&\beta\sum_{i=k+1}^n \{\sum_{t_i}p(t_i)\sum_{s_i}\frac{\delta p(s_i|t_i)}{\delta p(s_k|s_{k-1})}[log(p(s_i|t_i)) + 1] \\
    &-\sum_{s_i}\frac{\delta p(s_i)}{\delta p(s_k|s_{k-1})}[log(p(s_i)) + 1]) \} +\lambda(s_{k-1})
\end{split}
\end{equation}

Let us now derive for the expressions of the derivatives of the conditional probabilities.
Since each scene level is a strict compression of the previous level,
we have the Markov chain condition $\tasks_i \leftarrow \sgraph_0 \leftarrow \sgraph_1 \leftarrow \hdots \leftarrow \sgraph_n$ for all resolutions of the task description $\tasks_i$.
The conditional distributions for the first two levels are,
\begin{equation}
    p(s_1) = \sum_{s_0\in \sgraph_0} p(s_1 | s_0) p(s_0)
\end{equation}
\begin{equation}
    p(s_1|t_1) = \sum_{s_0\in \sgraph_0} p(s_1 | s_0) p(s_0|t_1)
\end{equation}
\begin{equation}
    p(s_2) = \sum_{s_1 \in \sgraph_1}\sum_{s_0 \in \sgraph_0} p(s_2 | s_1)p(s_1 | s_0) p(s_0)
\end{equation}
\begin{equation}
    p(s_2|t_2) = \sum_{s_1\in \sgraph_1} \sum_{s_0 \in \sgraph_0} p(s_2 | s_1) p(s_1|s_0) p(s_0|t_2)
\end{equation}
Generalized for level $n$ and some $k$ such that $n > k > 0$,
\begin{equation}\label{eq:psn}
\begin{split}
    &p(s_n) = \\
    &\sum_{s_k \in \sgraph_k}\sum_{s_{k-1} \in \sgraph_{k-1}} p(s_n | s_k) p(s_k | s_{k-1}) p(s_{k-1})
\end{split}
\end{equation}
\begin{equation}\label{psntn}
\begin{split}
    &p(s_n|t_n) = \\
    &\sum_{s_k \in \sgraph_k} \sum_{s_{k-1} \in \sgraph_{k-1}} p(s_n | s_k) p(s_k|s_{k-1})p(s_{k-1}|s_0)p(s_0|t_n)
\end{split}
\end{equation}

Taking the derivative of the conditional distributions with respect to $p(s_1|o) \hdots p(s_k|s_{k-1})$,
\begin{equation}
    \frac{\delta p(s_1)}{\delta p(s_1 | s_0)} = p(s_0)
\end{equation}
\begin{equation}
    \frac{\delta p(s_1|t_1)}{\delta p(s_1 | s_0)} = p(s_0|t_1)
\end{equation}
\begin{equation}\label{eq:dpsn}
    \frac{\delta p(s_n)}{\delta p(s_k | s_{k-1})} = p(s_n | s_k)p(s_{k-1})
\end{equation}
\begin{equation}\label{eq:dpsntn}
    \frac{\delta p(s_n|t_n)}{\delta p(s_k | s_{k-1})} = p(s_n|s_k)p(s_{k-1}|s_0)p(s_0|t_n)
\end{equation}

Substituting in the expressions \eqref{eq:dpsn} and \eqref{eq:dpsntn} into the derivative of the Lagrangian~\eqref{eq:deriv_lagrangian},
\begin{equation}\label{eq:k_level_halfway}
\begin{split}
    &\frac{\delta \mathcal{L}}{\delta p(s_k|s_{k-1})}\\
    =&p(s_{k-1}) \{ log(\frac{p(s_k|s_{k-1})}{p(s_k)}) \\
    +& \sum_{s_{k+1}}p(s_{k+1}|s_k) log(\frac{p(s_{k+1}|s_k)}{p(s_{k+1})}) \\
    +&\sum_{i=k+2}^n \{
    \sum_{s_{i-1}} p(s_{i-1}|s_k)\sum_{s_i} p(s_i | s_{i-1}) log(\frac{p(s_i|s_{i-1})}{p(s_i)}) \}\\
    &- (n-1) \\
    -&\beta\{\sum_{t_k}p(t_k)p(s_{k-1}|o)p(o|t_k)[log(p(s_k|t_k)) + 1] \\
    &-p(s_{k-1})[log(p(s_k)) + 1]) \}\\
    -&\beta\sum_{i=k+1}^n \{\\
    &\sum_{t_i}p(t_i)\sum_{s_i}p(s_i|s_k)p(s_{k-1}|s_0)p(s_0|t_i)[log(p(s_i|t_i)) + 1] \\
    &-\sum_{s_i}p(s_i|s_k)p(s_{k-1})[log(p(s_i)) + 1]) \} + \lambda(s_{k-1})
\end{split}
\end{equation}
We can rewrite $p(t_k)p(s_{k-1}|o)p(o|t_k)$ as
\begin{equation}
    p(t_k)p(s_{k-1}|t_k)= p(s_{k-1})p(t_k|s_{k-1})
\end{equation}
which allows us to simplify \eqref{eq:k_level_halfway} further as,
\begin{equation}\label{eq:k_level_deriv}
\begin{split}
    &\frac{\delta \mathcal{L}}{\delta p(s_k|s_{k-1})}\\
    =&p(s_{k-1}) \{ log(\frac{p(s_k|s_{k-1})}{p(s_k)}) \\
    +& \sum_{s_{k+1}}p(s_{k+1}|s_k) log(\frac{p(s_{k+1}|s_k)}{p(s_{k+1})}) \\
    +&\sum_{i=k+2}^n
    \sum_{s_{i-1}} p(s_{i-1}|s_k)\sum_{s_i} p(s_i | s_{i-1}) log(\frac{p(s_i|s_{i-1})}{p(s_i)}) \\
    &- (n-1) \\
    -&\beta\sum_{t_k}p(t_k|s_{k-1})log(\frac{p(s_k|t_k)}{p(s_k)}) \\
    -&\beta\sum_{i=k+1}^n
    \sum_{t_i}\sum_{s_i}p(s_i|s_k)p(t_i|s_{k-1})log(\frac{p(s_i|t_i)}{p(s_i)}) \} \\
    &+ \lambda(s_{k-1})
\end{split}
\end{equation}
Notice that the Kullback–Leibler divergence naturally emerges from the $\beta$ terms with some algebraic manipulation,
\begin{equation}
\begin{split}
    &p(t_i|s_{k-1})log(\frac{p(s_i|t_i)}{p(s_i)}) = -D_{KL}(p(t_i|s_{k-1}) || p(t_i|s_i)) \\
    &+ \sum_{t_i}p(t_i|s_{k-1})log(\frac{p(t_i|s_{k-1}}{p(t_i)})) \\
\end{split}
\end{equation}

Let us define $\tilde{\lambda}(s_{k-1})$ to group the terms that are not dependent on $s_k$,
\begin{equation}
\begin{split}
    &\tilde{\lambda}(s_{k-1}) = \frac{\lambda(s_{k-1})}{p(s_{k-1})} - (n-1) \\
    &+ \sum_{s_{k+1}}p(s_{k+1}|s_k) log(\frac{p(s_{k+1}|s_k)}{p(s_{k+1})}) \\
    &+\sum_{i=k+2}^n\sum_{s_{i-1}} p(s_{i-1}|s_k)\sum_{s_i} p(s_i | s_{i-1}) log(\frac{p(s_i|s_{i-1})}{p(s_i)}) \\
    &- \beta \sum_{t_i}p(t_i|s_{k-1})log(\frac{p(t_i|s_{k-1})}{p(t_i)})) \\
    &- \beta \sum_{i=k+1}^n \sum_{s_i}\sum_{t_i}p(t_i|s_{k-1})log(\frac{p(t_i|s_{k-1})}{p(t_i)})) \} \\
\end{split}
\end{equation}

Setting the derivative~\eqref{eq:k_level_deriv} to zero then gives us,
\begin{equation}
\begin{split}
    &0 = p(s_{k-1}) \{ log(\frac{p(s_k|s_{k-1})}{p(s_k)}) \\
    +&\beta D_{KL}(p(t_k|s_k) || p(t_k|s_{k-1})) \\
    +&\beta\sum_{i=k+1}^n\sum_{s_i}p(s_i|s_k)D_{KL}(p(t_i|s_i) || p(t_i|s_{k-1})) \\
    &+ \tilde{\lambda}(s_{k-1})\}
\end{split}
\end{equation}

Defining $\mathcal{Z} = \exp{[\tilde{\lambda}(s_{k-1})]}$, we have that 
\begin{equation}\label{eq:hib_sk}
\begin{split}
    p(s_k&|s_{k-1}) = \frac{p(s_k)}{\mathcal{Z}} \exp[-\beta D_{KL}(p(t_k|s_k) || p(t_k|s_{k-1})) \\
    &-\beta\sum_{i=k+1}^n\sum_{s_i}p(s_i|s_k)D_{KL}(p(t_i|s_i) || p(t_i|s_{k-1})) ] \\
\end{split}
\end{equation}
This corresponds to the iterative algorithm given in~\eqref{eq:hib_cases}.

\section{Tutorial on the Hierarchical Information Bottleneck}
\label{supp:hib_tutorial}

In this section, we provide a brief tutorial on \hib in the form of an easy example.
Given two tasks: $\mathcal{T} = \{ \Gamma, \Omega \}$,
three subtasks: $\mathcal{U} = \{ A, B, C \}$, and four items: $\mathcal{O} = \{ p, q, r, s \}$,
we want to assign each of $5$ primitives $\mathcal{X}$ to an item, each item to a subtask, and each subtask to a task,
and we are given the probability of how likely an observation might be that of an item $\prob{\mathcal{O}|\mathcal{X}}$,
the probability that an item might be relevant for a subtask $\prob{\mathcal{U}|\mathcal{O}}$,
and finally the probability that a subtask might be relevant for a task $\prob{\mathcal{T} | \mathcal{U}}$.
For this exercise, the conditional probabilities are given in Table~\ref{tab:conditional_prob_p_I_X}, Table~\ref{tab:conditional_prob_p_U_I},
and Table~\ref{tab:conditional_prob_p_T_U} respectively.
We treat each observation as identical and independent, so $\prob{\mathcal{X}}$ takes a uniform distribution $p(\mathcal{X} = x) = 0.2 \text{ } \forall x \in \mathcal{X}$. 
Note that the columns of conditional probability tables sum to one since these are probability mass functions.

Our goal is to find clusters $\mathcal{S_O}$, $\mathcal{S_U}$, and $\mathcal{S_T}$ where the assignments to the clusters are given by $\prob{\mathcal{S_O} | \mathcal{X}}$, $\prob{\mathcal{S_U} | \mathcal{S_O}}$, $\prob{\mathcal{S_T} | \mathcal{S_U}}$.
We initialize ($\tau = 0$) these conditional probabilities as Kronecker delta distributions and apply \hib.
We start from the object level to find $\mathbb{P}_{\tau = 1}(\mathcal{S_O} | \mathcal{X})$.
We start the iterative updates with the second and third equations of \eqref{eq:hib_cases}:
\begin{equation}
\small
p_{0}(s_o) = \sum_{x \in \mathcal{X}}p(x)p_0(s_o | x) \text{, } \forall s_o \in \mathcal{S_O}
\end{equation}

\begin{equation}
\small
p_{0}(x | s_o) = \frac{p(s_o|x)p(x)}{p(s_o)} \text{, }\forall (x, s_o) \in \mathcal{X} \times \mathcal{S_O}
\end{equation}

\begin{equation}
\small
p_{0}(o | s_o) = \sum_{x \in \mathcal{X}} p(o|x) p(x|s_o) \text{, }\forall (o, s_o) \in \mathcal{O} \times \mathcal{S_O}
\end{equation}

\begin{table}[h]
    \centering
    \begin{tabular}{c|ccccc}
        & \( x_1 \) & \( x_2 \) & \( x_3 \) & \( x_4 \) & \( x_5 \) \\
        \hline
        \( p \) & 0.7 & 0.6 & 0.1 & 0.1 & 0.1 \\
        \( q \) & 0.1 & 0.1 & 0.1 & 0.1 & 0.6\\
        \( r \) & 0.1 & 0.2 & 0.1 & 0.1 & 0.2 \\
        \( s \) & 0.1 & 0.1 & 0.7 & 0.7 & 0.1 \\
    \end{tabular}
    \caption{Conditional Probability Table $\prob{\mathcal{O}|\mathcal{X}}$}
    \label{tab:conditional_prob_p_I_X}
\end{table}

\begin{table}[h]
    \centering
    \begin{tabular}{c|cccc}
        & \( p \) & \( q \) & \( r \) & \( s \) \\
        \hline
        \( A \) & 0.8 & 0.2 & 0.1 & 0.1 \\
        \( B \) & 0.1 & 0.7 & 0.1 & 0.1 \\
        \( C \) & 0.1 & 0.1 & 0.8 & 0.8 \\
    \end{tabular}
    \caption{Conditional Probability Table $\prob{\mathcal{U}|\mathcal{O}}$}
    \label{tab:conditional_prob_p_U_I}
\end{table}

\begin{table}[h]
    \centering
    \begin{tabular}{c|ccc}
        & \( A \) & \( B \) & \( C \) \\
        \hline
        \( \Gamma \) & 0.9 & 0.1 & 0.2 \\
        \( \Omega \) & 0.1 & 0.9 & 0.8 \\
    \end{tabular}
    \caption{Conditional Probability Table $\prob{\mathcal{T}|\mathcal{U}}$}
    \label{tab:conditional_prob_p_T_U}
\end{table}

The expression to compute $\mathbb{P}_{\tau = 1}(\mathcal{S_O}|\mathcal{X})$ given in \eqref{eq:hib_sk} (also the first equation of \eqref{eq:hib_cases}) requires some manipulation:

\begin{equation}
\small
p_{0}(s_u | x) = \sum_{s_o \in \mathcal{S_O}} p_0(s_u|s_o) p_0(s_o|x) \text{, }\forall (s_u, x) \in \mathcal{S_U} \times \mathcal{X}
\end{equation}
\begin{equation}
\small
p_{0}(s_t | x) = \sum_{s_u \in \mathcal{S_U}} p_0(s_t|s_u) p_0(s_u|x) \text{, }\forall (s_t, x) \in \mathcal{S_T} \times \mathcal{X}
\end{equation}
\begin{equation}
\small
p(u | x) = \sum_{o \in \mathcal{O}} p(u|o) p(o|x) \text{, }\forall (u, x) \in \mathcal{U} \times \mathcal{X}
\end{equation}
\begin{equation}
\small
p_{0}(u | s_o) = \sum_{x \in \mathcal{X}} p(u|x) p_0(x|s_o) \text{, }\forall (u, s_o) \in \mathcal{U} \times \mathcal{S_O}
\end{equation}
\begin{equation}
\small
p_{0}(t | x) = \sum_{u \in \mathcal{U}} p_0(t|u) p_0(u|x) \text{, }\forall (t, x) \in \mathcal{T} \times \mathcal{X}
\end{equation}
\begin{equation}
\small
p_{0}(t | s_o) = \sum_{x \in \mathcal{X}} p_0(t|x) p_0(x|s_o) \text{, }\forall (t, s_o) \in \mathcal{T} \times \mathcal{S_O}
\end{equation}
\begin{equation}
\small
p_{0}(x | s_u) = \frac{p_0(s_u|x)p(x)}{p_0(s_u)} \text{, }\forall (x, s_u) \in \mathcal{X} \times \mathcal{S_U}
\end{equation}
\begin{equation}
\small
p_{0}(x | s_t) = \frac{p_0(s_t|x)p(x)}{p_0(s_t)} \text{, }\forall (x, s_t) \in \mathcal{X} \times \mathcal{S_T}
\end{equation}
\begin{equation}
\small
p_{0}(u | s_u) = \sum_{x \in \mathcal{X}} p(u|x) p_0(x|s_u) \text{, }\forall (u, s_u) \in \mathcal{U} \times \mathcal{S_U}
\end{equation}
\begin{equation}
\small
p_{0}(t | s_t) = \sum_{x \in \mathcal{X}} p(t|x) p_0(x|s_t) \text{, }\forall (t, s_t) \in \mathcal{T} \times \mathcal{S_T}
\end{equation}

Finally, we can plug in the values to \eqref{eq:hib_sk} to obtain
\begin{equation}\label{eq:hib_sk}
\small
\begin{split}
    &p_1(s_o|x) = \frac{p(s_o)}{\mathcal{Z}} \exp[-\beta D_{KL}(\prob{\mathcal{O}|\mathcal{S_O} = s_o} || \prob{\mathcal{O}|\mathcal{X} = x}) \\
    &-\beta \sum_{s_u \in \mathcal{S_U}} p(s_u|x) D_{KL}(\prob{\mathcal{U}|\mathcal{S_U} = s_u} || \prob{\mathcal{U}|\mathcal{X} = x}) \\
    &-\beta \sum_{s_t \in \mathcal{S_T}} p(s_t|x) D_{KL}(\prob{\mathcal{T}|\mathcal{S_T} = s_t} || \prob{\mathcal{T}|\mathcal{X} = x})]\\
\end{split}
\end{equation}

\begin{table}[h]
    \centering
    \begin{tabular}{c|ccccc}
        & \( x_1 \) & \( x_2 \) & \( x_3 \) & \( x_4 \) & \( x_5 \) \\
        \hline
        \( s_1 \) & 1.0 & 0.0 & 0.0 & 0.0 & 0.0 \\
        \( s_2 \) & 0.0 & 1.0 & 0.0 & 0.0 & 0.0 \\
        \( s_3 \) & 0.0 & 0.0 & 0.5 & 0.5 & 0.0 \\
        \( s_4 \) & 0.0 & 0.0 & 0.5 & 0.5 & 0.0 \\
        \( s_5 \) & 0.0 & 0.0 & 0.0 & 0.0 & 1.0 \\
    \end{tabular}
    \caption{Conditional Probability Table $\mathbb{P}_{\tau = 1}(\mathcal{S_O}|\mathcal{X})$}
    \label{tab:conditional_prob_p0_S_X}
\end{table}
\begin{table}[h]
    \centering
    \begin{tabular}{c|ccccc}
        & \( s_1 \) & \( s_2 \) & \( s_3 \) & \( s_4 \) & \( s_5 \) \\
        \hline
        \( p \) & 0.7 & 0.6 & 0.1 & 0.1 & 0.1 \\
        \( q \) & 0.1 & 0.1 & 0.1 & 0.1 & 0.6 \\
        \( r \) & 0.1 & 0.2 & 0.1 & 0.1 & 0.2 \\
        \( s \) & 0.1 & 0.1 & 0.7 & 0.7 & 0.1 \\
    \end{tabular}
    \caption{Conditional Probability Table $\mathbb{P}_{\tau = 1}(\mathcal{O}|\mathcal{S_O})$}
    \label{tab:conditional_prob_p0_O_SO}
\end{table}

Setting $\beta = 100$, we obtain an updated $\mathbb{P}_{\tau = 1}(\mathcal{S_O}|\mathcal{X})$ given in Table~\ref{tab:conditional_prob_p0_S_X}.

This conditional probability informs us of a \emph{soft cluster} map to group the primitives to objects.
In this case, $x_1$, $x_2$, $x_5$ each corresponds to an object and $x_3$, $x_4$ are grouped together as primitives of the same object.
Furthermore, we can \emph{label} these clusters by taking the argmax of $\prob{\mathcal{O}|\mathcal{S_O}}$, which is shown in Tab.~\ref{tab:conditional_prob_p0_O_SO} and can be obtained by manipulating the probability as follows,
\begin{equation}
\small
p_{1}(o | s_o) = \sum_{x \in \mathcal{X}} p(o|x) p_1(x|s_o) \text{, }\forall (o, s_o) \in \mathcal{O} \times \mathcal{S_O}
\end{equation}
To summarize, after this first iteration of just the object layer, 4 objects are obtained, an object with label $p$ consisting of the primitive $x_1$, an object with label $p$ consisting of the primitive $x_2$, an object with label $s$ consisting of the primitives $x_3$ and $x_4$, and an object with label $q$ consisting of the primitive $x_5$.

\begin{figure}[ht]
\centering
\includegraphics[width=0.4\columnwidth]{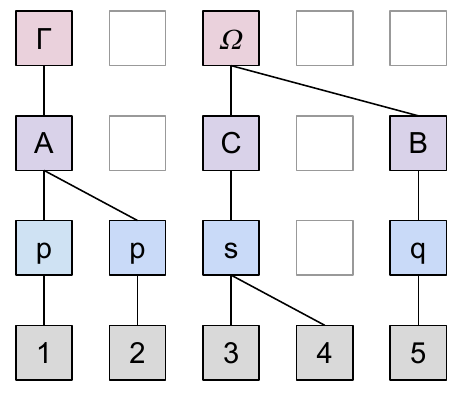}
\caption{Final hierarchy from HIB for tutorial example.}
\label{fig:tutorial}
\end{figure}

We repeat this for all three layers and for $n$ iterations until convergence. 
The final $\mathbb{P}_{\tau = n}(\mathcal{S_O} | \mathcal{X})$, $\mathbb{P}_{\tau = n}(\mathcal{S_U} | \mathcal{S_O})$, $\mathbb{P}_{\tau = n}(\mathcal{S_T} | \mathcal{S_U})$ is given in Table~\ref{tab:conditional_prob_p_SO_X}, Table~\ref{tab:conditional_prob_p_SU_SO},
and Table~\ref{tab:conditional_prob_p_ST_SU} respectively with associated
final $\mathbb{P}_{\tau = n}(\mathcal{O} | \mathcal{S_O})$, $\mathbb{P}_{\tau = n}(\mathcal{U} | \mathcal{S_U})$, $\mathbb{P}_{\tau = n}(\mathcal{T} | \mathcal{S_T})$ given in Table~\ref{tab:conditional_prob_p_O_SO}, Table~\ref{tab:conditional_prob_p_U_SU},
and Table~\ref{tab:conditional_prob_p_T_ST}. 
The combined gives us a hierarchy where task $\Gamma$ consists of subtask $A$, which consists of 2 objects both with label $p$ that came from two different primitives $x_1$ and $x_2$,
and task $\Omega$ consists of 2 subtasks $B$ and $C$,
subtask $B$ consists of an object with label $q$ from primitive $x_5$
and subtask $C$ consists of an object with label $s$
from primitives $x_3$ and $x_4$.
This is visually shown in Fig.~\ref{fig:tutorial},
where the edges represent the cluster assignment from taking the argmax of the conditional probabilities.

\begin{table}[h]
    \centering
    \begin{tabular}{c|ccccc}
        & \( x_1 \) & \( x_2 \) & \( x_3 \) & \( x_4 \) & \( x_5 \) \\
        \hline
        \( s_1 \) & 0.99 & 0.03 & 0.0 & 0.0 & 0.0 \\
        \( s_2 \) & 0.01 & 0.97 & 0.0 & 0.0 & 0.0 \\
        \( s_3 \) & 0.0 & 0.0 & 0.5 & 0.5 & 0.0 \\
        \( s_4 \) & 0.0 & 0.0 & 0.5 & 0.5 & 0.0 \\
        \( s_5 \) & 0.0 & 0.0 & 0.0 & 0.0 & 1.0 \\
    \end{tabular}
    \caption{Conditional Probability Table final $\mathbb{P}_{\tau = n}(\mathcal{S_O} | \mathcal{X})$}
    \label{tab:conditional_prob_p_SO_X}
\end{table}

\begin{table}[h]
    \centering
    \begin{tabular}{c|ccccc}
        & \( s_1 \) & \( s_2 \) & \( s_3 \) & \( s_4 \) & \( s_5 \) \\
        \hline
        \( s_1 \) & 0.51 & 0.51 & 0.0 & 0.0 & 0.0 \\
        \( s_2 \) & 0.49 & 0.49 & 0.0 & 0.0 & 0.0 \\
        \( s_3 \) & 0.0 & 0.0 & 0.5 & 0.5 & 0.0 \\
        \( s_4 \) & 0.0 & 0.0 & 0.5 & 0.5 & 0.0 \\
        \( s_5 \) & 0.0 & 0.0 & 0.0 & 0.0 & 1.0 \\
    \end{tabular}
    \caption{Conditional Probability Table final $\mathbb{P}_{\tau = n}(\mathcal{S_U} | \mathcal{S_O})$}
    \label{tab:conditional_prob_p_SU_SO}
\end{table}

\begin{table}[h]
    \centering
    \begin{tabular}{c|ccccc}
        & \( s_1 \) & \( s_2 \) & \( s_3 \) & \( s_4 \) & \( s_5 \) \\
        \hline
        \( s_1 \) & 0.51 & 0.51 & 0.0 & 0.0 & 0.0 \\
        \( s_2 \) & 0.49 & 0.49 & 0.0 & 0.0 & 0.0 \\
        \( s_3 \) & 0.0 & 0.0 & 0.33 & 0.33 & 0.33 \\
        \( s_4 \) & 0.0 & 0.0 & 0.33 & 0.33 & 0.33 \\
        \( s_5 \) & 0.0 & 0.0 & 0.33 & 0.33 & 0.33 \\
    \end{tabular}
    \caption{Conditional Probability Table final $\mathbb{P}_{\tau = n}(\mathcal{S_T} | \mathcal{S_U})$}
    \label{tab:conditional_prob_p_ST_SU}
\end{table}

\begin{table}[h]
    \centering
    \begin{tabular}{c|ccccc}
        & \( s_1 \) & \( s_2 \) & \( s_3 \) & \( s_4 \) & \( s_5 \) \\
        \hline
        \( p \) & 0.7 & 0.6 & 0.1 & 0.1 & 0.1 \\
        \( q \) & 0.1 & 0.1 & 0.1 & 0.1 & 0.6 \\
        \( r \) & 0.1 & 0.2 & 0.1 & 0.1 & 0.2 \\
        \( s \) & 0.1 & 0.1 & 0.7 & 0.7 & 0.1 \\
    \end{tabular}
    \caption{Conditional Probability Table $\mathbb{P}_{\tau = n}(\mathcal{O}|\mathcal{S_O})$}
    \label{tab:conditional_prob_p_O_SO}
\end{table}

\begin{table}[h]
    \centering
    \begin{tabular}{c|ccccc}
        & \( s_1 \) & \( s_2 \) & \( s_3 \) & \( s_4 \) & \( s_5 \) \\
        \hline
        \( A \) & 0.57 & 0.57 & 0.18 & 0.18 & 0.23 \\
        \( B \) & 0.16 & 0.16 & 0.16 & 0.16 & 0.46 \\
        \( C \) & 0.27 & 0.27 & 0.66 & 0.66 & 0.31 \\
    \end{tabular}
    \caption{Conditional Probability Table $\mathbb{P}_{\tau = n}(\mathcal{U}|\mathcal{S_U})$}
    \label{tab:conditional_prob_p_U_SU}
\end{table}

\begin{table}[h]
    \centering
    \begin{tabular}{c|ccccc}
        & \( s_1 \) & \( s_2 \) & \( s_3 \) & \( s_4 \) & \( s_5 \) \\
        \hline
        \( \Gamma \) & 0.58 & 0.58 & 0.31 & 0.31 & 0.31 \\
        \( \Omega \) & 0.42 & 0.42 & 0.69 & 0.69 & 0.69 \\
    \end{tabular}
    \caption{Conditional Probability Table $\mathbb{P}_{\tau = n}(\mathcal{T}|\mathcal{S_T})$}
    \label{tab:conditional_prob_p_T_ST}
\end{table}

\end{document}